\pdfoutput=1

\documentclass[11pt]{article}

\usepackage[]{acl}

\usepackage{times}
\usepackage{latexsym}

\usepackage{graphicx}
\usepackage[T1]{fontenc}
\usepackage{hhline,multirow}
\usepackage{longtable}
\usepackage{amssymb}
\usepackage{amsmath}
\usepackage{paralist}
\usepackage{fontawesome5}

\usepackage[utf8]{inputenc}
\usepackage{amsmath}
\usepackage{amsfonts}
\usepackage{hhline}
\usepackage{multirow}
\usepackage{makecell}
\usepackage{rotating}
\usepackage{makecell}
\usepackage{lipsum}
\usepackage{hyperref}

\usepackage{microtype}

\title{MedLogic-AQA: Enhancing Medical Question Answering with Abstractive Models Focusing on Logical Structures}

\author{
    Aizan Zafar\textsuperscript{1*} \quad Kshitij Mishra\textsuperscript{1*} \quad Asif Ekbal\textsuperscript{2} \\
    \textsuperscript{1}Department of Computer Science and Engineering, Indian Institute of Technology Patna, India \\
    \textsuperscript{2}School of AI and Data Science, Indian Institute of Technology Jodhpur, India \\
    \texttt{aizanzafar@gmail.com, mishra.kshitij07@gmail.com, asif.ekbal@gmail.com}
}

\begin{document}
\maketitle

\begingroup\renewcommand\thefootnote{\fnsymbol{footnote}}
\footnotetext[1]{Equal contribution.}
\endgroup
\begin{abstract}
In Medical question-answering (QA) tasks, the need for effective systems is pivotal in delivering accurate responses to intricate medical queries. However, existing approaches often struggle to grasp the intricate logical structures and relationships inherent in medical contexts, thus limiting their capacity to furnish precise and nuanced answers. In this work, we address this gap by proposing a novel Abstractive QA system \textsc{MedLogic-AQA} that harnesses First Order Logic (FOL) based rules extracted from both context and questions to generate well-grounded answers. Through initial experimentation, we identified six pertinent first-order logical rules, which were then used to train a Logic-Understanding (LU) model capable of generating logical triples for a given context, question, and answer. These logic triples are then integrated into the training of \textsc{MedLogic-AQA}, enabling effective and coherent reasoning during answer generation. This distinctive fusion of logical reasoning with abstractive QA equips our system to produce answers that are logically sound, relevant, and engaging. 
Evaluation with respect to both automated and human-based demonstrates the robustness of \textsc{MedLogic-AQA} against strong baselines. Through empirical assessments and case studies, we validate the efficacy of \textsc{MedLogic-AQA} in elevating the quality and comprehensiveness of answers in terms of reasoning as well as informativeness  
\footnote{Code: \hyperlink{https://github.com/aizanzafar/MedLogicAQA}{https://github.com/aizanzafar/MedLogicAQA}}.

\end{abstract}

\section{Introduction}
In recent years, the demand for effective question-answering (QA) systems in the field of medicine has surged, driven by the need to provide accurate and informative responses to complex medical inquiries \cite{shickel2018deep}. With the proliferation of medical data and the increasing reliance on digital platforms for healthcare information, the development of robust QA systems has become imperative to support medical professionals and patients alike \cite{pons2016natural}.

Existing abstractive question-answering (AQA) approaches in medicine face significant challenges in capturing the intricate logical structures and relationships inherent in medical contexts \cite{minsky1975framework, zhu2020knowledge, zafar2023ki}. This leads to sub-optimal outcomes \cite{rajpurkar2018squad}, i.e. limiting the AQA systems to furnish precise and nuanced answers to medical queries necessitating logical reasoning \cite{cappanera2023logic}. Through the integration of logical reasoning in AQA systems, they can navigate the complexities of medical data and provide reasoned, coherent, and informative answers to medical queries \cite{ratner2017snorkel, choi2017doctor}. 

\begin{figure*}
\includegraphics[width=16 cm,height=8cm]{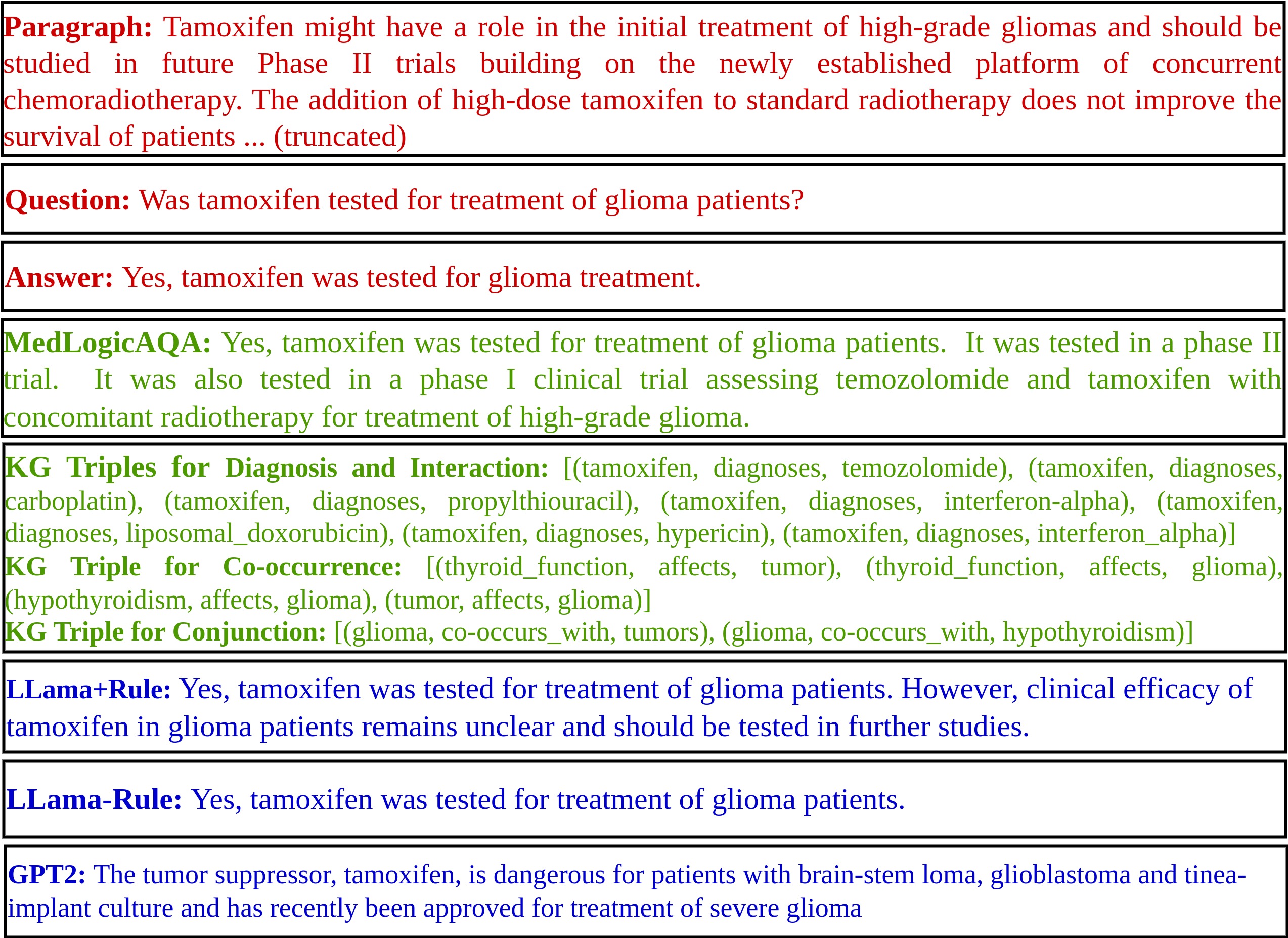}
    \caption{Illustration of Responses Generated by \textsc{MedLogic-AQA}: Demonstrating the approach's utilization of background knowledge and first-order logic-based rules to provide comprehensive answers to medical queries, exemplifying its logical reasoning capabilities"}
    \label{fig:example}
\end{figure*}

Therefore, to address the limitations of the existing approaches, We propose a novel abstractive QA system, \textsc{MedLogic-AQA}. Central to our approach is the conceptualization of the logical structure of the context as a graph. We achieve this by employing six carefully chosen First Order Logical (FOL) rules, with nodes representing entities and edges encapsulating their logical relationships. This structured representation facilitates a more nuanced understanding of the context, enabling us to navigate intricate logical dependencies effectively. Utilizing this graph-based information as output and with the context, question, and answer as input, we initially train a Logic-Understanding (LU) model using LLAMA2 \cite{touvron2023llama}. Subsequently, we fine-tune the LU model with the input of context, question, and the desired output answer. This fine-tuning process emphasizes logical coherence and contextual relevance, enhancing the generation of answers. This logic-based representation offers a flexible and scalable framework for integrating logical rules, making it applicable across diverse domains and datasets.

An example of \textsc{MedLogic-AQA} is shown in Figure \ref{fig:example}. It can be seen that \textsc{MedLogic-AQA} first establishes the background knowledge of different entities involved with the user to provide the answer for a better understanding. To perform reasoning, \textsc{MedLogic-AQA} leverages first-order logic-based rules extracted from both context and questions to generate well-grounded answers that encapsulate the underlying logical reasoning behind medical concepts and relationships \cite{wang2021logic}. Our \textit{key} contributions can be summarized as follows:
\begin{enumerate}
    \item Proposed an effective neuro-symbolic approach that leverages first-order logic reasoning in a neural network framework for Medical Abstractive Question Answering System \textsc{MedLogic-AQA}. 
    \item Develop a Logic Understanding model that generates Logic triples without the need for any traditional graph-based method.
    \item Through a series of empirical evaluation and case studies, we demonstrate the efficacy of \textsc{MedLogic-AQA} in elevating the quality and comprehensiveness of answers provided in terms of reasoning as well as informativeness.
\end{enumerate}

\section{Related Work}

Abstractive Question Answering (AQA) has witnessed substantial research efforts, with several approaches aiming to enhance the generation of contextually relevant and coherent answers \cite{fan2019eli5,krishna2021hurdles,pal2022parameter}. The pursuit of effective question-answering (QA) systems in medical domain has garnered considerable attention in the recent years \cite{shickel2018deep}. This surge in interest stems from the critical necessity of furnishing accurate and informative responses to intricate medical inquiries amidst the proliferation of medical data and the increasing reliance on digital platforms for healthcare information \cite{pons2016natural}. Existing literature primarily falls into two categories: methods leveraging neural networks and those incorporating logical reasoning.

\textbf{Neural Network-based Approaches:}
Early endeavors in AQA predominantly focused on neural network-based models, often employing recurrent neural networks (RNNs) and later transitioning to attention mechanisms and transformers \cite{vaswani2017attention}. Notable works include the introduction of sequence-to-sequence models \cite{sutskever2014sequence}. While these methods demonstrated promising results, they struggled to capture intricate logical structures and dependencies within the context, limiting their ability to handle complex queries that require nuanced reasoning.

\textbf{Logical Reasoning in Question Answering:}
Recognizing the limitations of neural network-centric approaches, researchers delved into incorporating logical reasoning to imbue AQA systems with enhanced inferential capabilities \cite{moldovan2003cogex,asai2020logic,li2019augmenting}. Early attempts utilized knowledge graphs and semantic parsing to introduce explicit logical structures \cite{berant2014semantic}. However, these methods faced challenges in scalability and were often domain-specific.

\textbf{Graph-Based Representations:}
Recent advancements in graph-based representations \cite{lin2022effectively,fouladvand2023graph} have offered a more versatile and scalable approach to capturing logical relationships within text. Graph neural networks (GNNs) have shown promise in modeling dependencies and hierarchies in various natural language processing tasks \cite{zhang2020graph,huai2023spatial,amador2023geni}. However, the application of GNNs in AQA \cite{zafar2023ki} has been limited, and their efficacy in handling logical rules derived from the context remains an under-explored area.


In healthcare, several attempts have been made to develop persuasive \cite{mishra2022please, samad2022empathetic} and counseling conversation systems \cite{mishra2023help, mishra2023pal, priya2023partner, mishra2023therapist}. However, these systems primarily focus on enhancing meta-communicative aspects, such as politeness, empathy, and personalization, rather than generating context-sensitive responses. Specifically, within the domain of medical care, while there has been work in the field of medicine, current QA approaches face significant challenges in capturing the complex logical structures and relationships inherent in medical contexts \cite{zhu2020knowledge,varshney2023knowledge,zafar2024my, zafar2024kimedqa, varshney2022cdialog}. The inability to effectively discern intricate logical patterns within medical data often leads to sub-optimal results, impacting both the accuracy and relevance of the answers provided \cite{leaman2015challenges}. These limitations hinder the ability of QA systems to offer precise and nuanced responses to medical queries that demand logical reasoning \cite{huth2004logic}. \citet{wang2021logic} proposed a logic-based approach that leverages first-order logic rules extracted from both the context and questions to generate well-grounded answers, incorporating the underlying logical reasoning embedded within medical concepts and relationships.


This work bridges the gap between neural network-based AQA models and logical reasoning by proposing a novel framework that leverages first-order logic-based rules extracted from the context, represented as a graph. Our approach draws inspiration from Minsky's seminal work on knowledge representation \cite{minsky1975framework}, aiming to integrate explicit logical structures into the AQA process. Additionally, the attention mechanism proposed by \citet{vaswani2017attention} serves as a cornerstone in our approach, facilitating the nuanced integration of logical rules into the abstractive question-answering paradigm. Unlike previous works, our method focuses on the extraction of logical rules directly from the context, enabling a more dynamic and context-aware system.

\section{Methodology}
The proposed system \textsc{MedLogic-AQA} involves two components, \textit{viz.} (i.) \textit{Logic Understanding} module -  responsible for infusing logical rules into the model's decision-making process. It plays a critical role in enhancing the model's reasoning capabilities, making it adept at understanding complex relationships and dependencies within the data. (ii.) \textit{MedAQA} module -  this step utilizes LU's logical reasoning capabilities to refine the model's understanding of complex dependencies to generate logically correct and contextually relevant answers as per first-order logic rules. The two-stage fine-tuning approach is detailed in Section \ref{luaqa} of the appendix. 

\subsection{Logic Understanding Module}
\textbf{Medical Knowledge Graph Creation:} We construct a self-built knowledge graph using Quick-UMLS \cite{soldaini2016quickumls}, which is based on the UMLS \cite{bodenreider2004unified}. \textit{Knowledge Construction:} To construct knowledge graph (KG) triples, each context is processed through the UMLS \cite{bodenreider2004unified} to generate a smaller and more pertinent KG. \textit{Medical Entity Extraction:} We identify medical entities from each context by employing the Metathesaurus. Each distinct concept found in the UMLS is represented as a node in our knowledge graph. \textit{Relation Extraction:} Relations within our knowledge graph are sourced from both the \textit{Metathesaurus} and the Semantic Network of UMLS. \textit{Graph Construction:} Using the extracted relations from both sources, we establish connections between the filtered medical concepts retrieved from UMLS. These steps result in a Medical Knowledge Graph (MKG) that enriches our understanding of medical concepts and their relationships for the given question $q_i$ and context $c_i$.

\begin{figure*}
    \centering
    \includegraphics[width=16 cm, height=10cm]{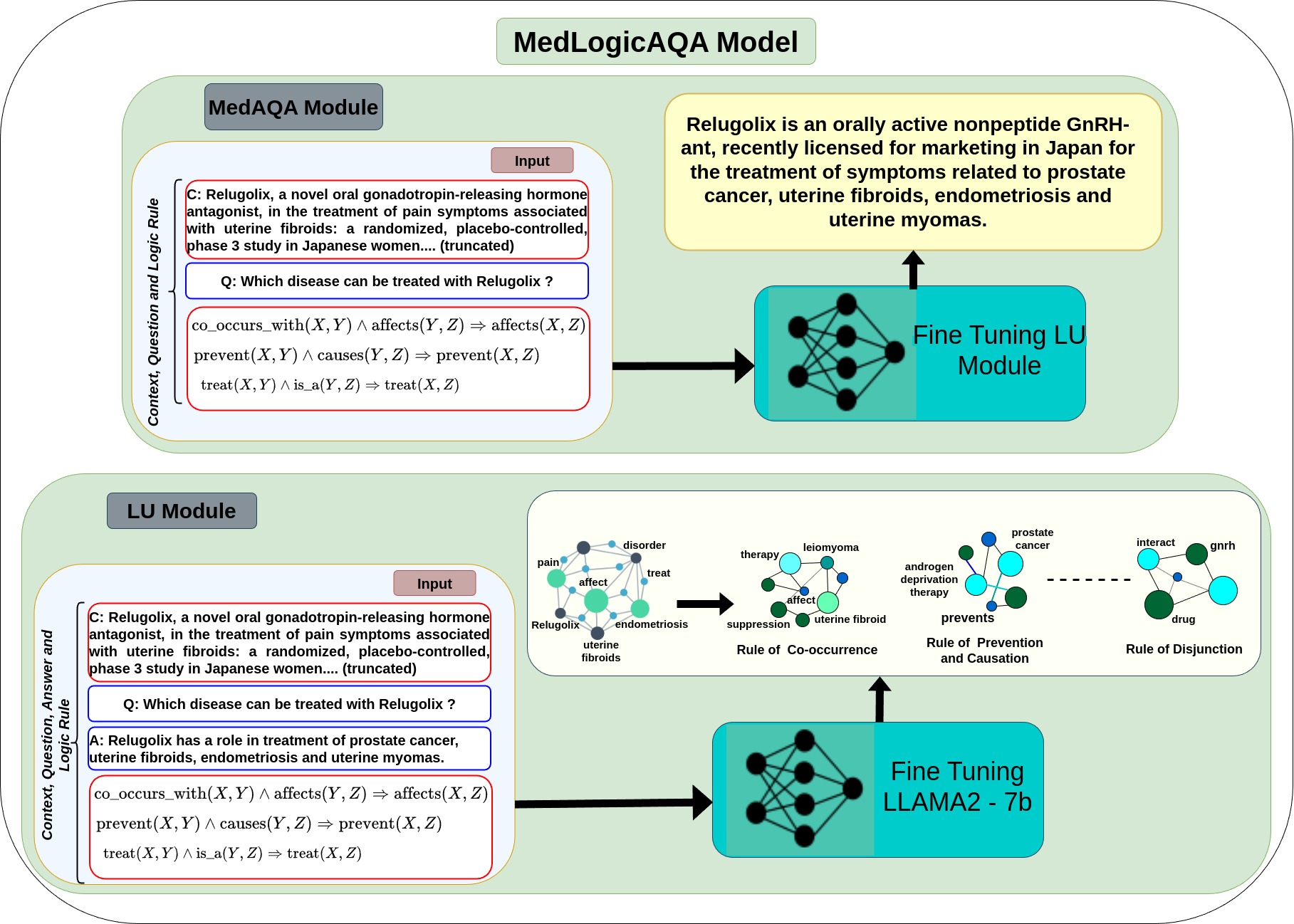}
    \caption{Illustration of architecture of the {\fontfamily{put}\selectfont \textbf{MedLogic-AQA}} system. The Logic Understanding (LU) Module comprises several components: Context, Question, Logical Rule, and Answer. These components are input to the LLama2-7B model to generate logical knowledge triples. Subsequently, the LU module is fine-tuned using the context, logical rule, and question to generate the final answer. }
    \label{fig:architecture}
\end{figure*}
\noindent \textbf{Logic Rule Injection:} 
After going through $k$ number of FOL-based rules, we finalized six rules which were relevant and were able to serve as essential knowledge for enhancing the model's reasoning and inference capabilities. Additional information about the derivation of logical rules can be found in the Appendix \ref{apx:logic_rule}.

\noindent 
\begin{enumerate}
    \item \textbf{Rule of Co-occurrence:} If entity X co-occurs with entity Y, and Y affects entity Z, then entity X also affects entity Z.
    \begin{multline}
    \text{co\_occurs\_with}(X, Y) \land \text{affects}(Y, Z) \Rightarrow \\
    \text{affects}(X, Z)
    \end{multline}

    \item \textbf{Rule of Prevention and Causation:} If intervention X prevents event Y, and Y causes event Z, it can be inferred that X can also prevent Z.
    \begin{multline}
    \text{prevent}(X, Y) \land \text{causes}(Y, Z) \Rightarrow \\
    \text{prevent}(X, Z)
    \end{multline}

    \item \textbf{Rule of Treatment and Classification:} If treatment X is effective for condition Y, and Y is a type of condition Z, then X can also be used to treat Z.
    \begin{multline}
    \text{treat}(X, Y) \land \text{is\_a}(Y, Z) \Rightarrow \\
    \text{treat}(X, Z)
    \end{multline}

    \item \textbf{Rule of Diagnosis and Interaction:} If entity X is diagnosed with condition Y, and X interacts with entity Z, it suggests that Z can be used for the diagnosis of Y.
    \begin{multline}
    \text{diagnosis}(X, Y) \land \text{interacts\_with}(X, Z) \Rightarrow \\
    \text{diagnosis}(Z, Y)
    \end{multline}

    \item \textbf{Rule of Conjunction:} If entity X co-occurs with entity Y and X affects entity Z, it implies that Y and Z also co-occur.
    \begin{multline}
    \text{co\_occurs\_with}(X, Y) \land \text{affects}(X, Z) \Rightarrow \\
    \text{co\_occurs\_with}(Y, Z)
    \end{multline}

    \item \textbf{Rule of Disjunction:} If either entity X prevents Y or Y causes Z, then it can be inferred that either X prevents Z or X causes Z.
    \begin{multline}
    \text{prevent}(X, Y) \lor \text{causes}(Y, Z) \Rightarrow \\
    (\text{prevent}(X, Z) \lor \text{causes}(X, Z))
    \end{multline}
\end{enumerate}

The logical rules described above are integrated with the MKG triples. Triples that meet the criteria of a given first-order logical rule $R_k$ are selected as outputs, where $X$ and $Y$ represent the head and tail of a triple, respectively, and the relationship is a function, such as "affects" or "causes" applied to these triples. Consequently, each rule produces a set of logic-injected triples. These triples obtained from each rule are aggregated to obtain a logic-injected knowledge graph.

\noindent \textbf{Logic Graph Learning:}
We fine-tune LLama2-7b-hf \cite{touvron2023llama} to obtain the Logic Understanding (LU) model. It learns the logic graphs for the given input $x = [q_i+c_i+a_i+R_i]$ and output $y = lt$: where, $q_i$, $c_i$, $a_i$, and $R_i$ is $i^{th}$ represent the question, context, answer, and set of six predefined logical rules, respectively; $lt$ denotes the obtained logical triples. $LU_\theta$ is the approximated probability distribution $p_{\theta}(y | x)$ on all $N$ instances of the given input-output $(x_i, y_i)$ pairs, where $0 \leq i <n$. 
\begin{equation}{p_{\theta}(y | x) = \text{softmax}(f_{\theta}(x, y))}\end{equation} 
$f_{\theta}(x, y)$ is the output of the LLama2-7b-hf when fine-tuning. $p_{\theta}(y | x)$ is the probability of generating $y$ given input $x$.
\begin{equation} \mathcal{L}(\theta) = -\frac{1}{N} \sum_{i=1}^{N} \log p_{\theta}(y_i | x_i) \end{equation}
$\mathcal{L}(\theta)$ is the computed cross-entropy loss on $N$ instances. The parameters are updated as follows:
 \begin{equation} \theta_{t+1} = \theta_{t} - \alpha \nabla_{\theta} \mathcal{L}(\theta) \end{equation}
This training process enables the model to learn the logic embedded in the context in the form of rules.

\subsection{\textsc{MedLogic-AQA}}
To obtain \textsc{MedLogic-AQA}, LU is further fine-tuned with input $x = [q_i+c_i+R_i]$, and output $y = a_i$. Building upon the knowledge acquired by LU, this step utilizes its logical reasoning capabilities to refine the model's understanding of complex dependencies. This ensures the generation of logically correct and contextually relevant answers based on the learned rules. The inclusion of logical rules in both fine-tuning stages contributes to making the model more context-aware and adaptable to the intricacies of medical queries. The overall architecture of the proposed system can be seen in Figure \ref{fig:architecture}. 

\section{Dataset}
Our experiments are conducted on two benchmark datasets: MASH-QA \cite{zhu2020question} and the BioASQ Task 10b Phase B (QA task) dataset \cite{nentidis2022overview}.

The BioASQ Task 10b Phase B \cite{nentidis2022overview} dataset is meticulously designed for biomedical QA, encompassing tasks like biomedical semantic indexing and QA, with a specific emphasis on the QA task. On the other hand, the MASH-QA \cite{zhu2020question} dataset consists of consumer healthcare questions extracted from WebMD, covering diverse healthcare sectors and addressing common healthcare concerns. With approximately 25K question-answer pairs, it is the largest dataset available in the medical domain. For detailed dataset statistics and pre-processing information, please refer to the Appendix \ref{apx:dataset}.

\section{Experiments}
\subsection{Baselines}
We compare the proposed \textsc{MedLogic-AQA} to seven strong baselines, BART \cite{lewis2020bart}, GPT2 \cite{radford2019language}, BioGPT \cite{luo2022biogpt}, BioMistral-7B \cite{labrak2024biomistral}, BioMedGPT-LM-7B \cite{luo2023biomedgpt}, LLama2-Rule - Fine-tuning LLamA2 \cite{touvron2023llama} considering input: $x = [q_i+c_i]$, and output $y = a_i$, LLama2+Rule -  Fine-tuning LLamA2 \cite{touvron2023llama} considering input: $x = [q_i+c_i+R_i]$, and output $y = a_i$. Additional information about baselines can be found in Appendix \ref{apx:baseline-details}.

\subsection{Implementation Details}
We implement all the models on a train:test split of 80:20. For all the models, we used random\_seed=40, learning rate = 1e-5, dropout = 0.2, Adam optimizer \cite{loshchilov2018decoupled}, and n\_epochs = 15. The implementation utilized the A100-PCIE-40GB with CUDA version 11.2 for GPU acceleration. Each training epoch lasted approximately 4.5 hours. Additional information about hyperparameters can be found in the Appendix \ref{sec:hyperparameters}.

\subsection{Evaluation Metrics}
\paragraph{\textbf{Automatic Evaluation:}} All the models are evaluated on the test set, using the standard metrics: BLEU score \cite{papineni2002bleu} - checks word overlap between predicted and ground truth responses, ROUGE-L \cite{lin-2004-rouge} assesses the longest matching word sequence, METEOR \cite{banerjee2005meteor}, Medical Entity F1-score \footnote{https://github.com/facebookresearch/ParlAI/parlai/metrics.py} computed by comparing predicted and ground truth sentences, Embedding-based metrics (i.e. Embedding Average metric)\footnote{https://github.com/Maluuba/nlg-eval} \cite{liu2016not} and A-LEN gives the average number of tokens in the generated answer.

\paragraph{\textbf{Human Evaluation:}} 
Automated metrics alone cannot fully capture critical aspects, such as the adequacy of logical reasoning, contextual consistency, or response accuracy. Therefore, a human evaluation was conducted on the generated answers from all models. To evaluate the quality of responses, we selected 120 generated answers along with their corresponding questions, contexts, and ground-truth answers from the BioASQ dataset. Five human evaluators were recruited to assess answer quality across four dimensions: \textit{Adequacy}, which examines whether the response is relevant and meaningful; \textit{Fluency}, which measures grammatical correctness; \textit{Logical Reasoning}, which evaluates the coherence and correctness of reasoning based on the provided context and question; and \textit{Contextual Consistency}, which checks whether the answer aligns with the given context.

All evaluators hold postgraduate qualifications in linguistics and possess substantial experience in related evaluation tasks. The models were rated on a 5-point Likert scale, with 1 representing the lowest performance and 5 representing the highest, across all metrics. The inter-evaluator agreement scores \cite{cohen1960coefficient} for \textit{Adequacy}, \textit{Fluency}, \textit{Logical Reasoning}, and \textit{Contextual Consistency} were 81.3\%, 85.6\%, 80.1\%, and 83.5\%, respectively, confirming substantial agreement. For more detailed information, please refer to the Appendix \ref{apx:human_eval}

\begin{table*}[hbt!]\footnotesize
\centering
\begin{tabular}{l|ccccccc}
\hline
\textbf{Models} &  \textbf{Medical Entity F1\% } &  \textbf{BLEU} &  \textbf{ROUGE-L } & \textbf{METEOR} & \textbf{Embedding Average} & \textbf{A-LEN} \\
\hline
GPT-2 & 8.52 & 0.0094 &  0.0678 & 0.1087 & 0.708 & 20.66   \\
BART  & 21.68 & 0.209 &  0.2468 & 0.4083 & 0.779 & 37.85   \\
BioGPT  & 10.96 & 0.0294 &  0.1074 & 0.2166 & 0.732 & 24.85   \\
BioMistral  & 19.19 & 0.2053 &  0.2599 & 0.4153 & 0.780 & 55.20   \\
BioMedGPT-LM  & 15.53 & 0.1715 &  0.2314 & 0.3549 & 0.778 & 32.85   \\
\text{LLama2-Rule}  & 24.88 & 0.2309 &  0.2615 & 0.4220 & 0.782 & 48.28   \\
\text{LLama2+Rule}  & 25.12 & 0.2476 & 0.2626 & 0.4248 & 0.821 & 70.57   \\
\hline
\textsc{\textbf{MedLogic-AQA}} & \textbf{38.47} & \textbf{0.2729} &  \textbf{0.2768} & \textbf{0.4383} & \textbf{0.838} & {53.71}   \\
\hline
\end{tabular}
\caption{\label{tab:auto_results_bioasq}
{Automatic evaluation results of BioASQ dataset. Here, LLama2-Rule represents fine-tuned LLama2-7b model only on given context and question to generate answer, while LLama2+Rule represents fine-tuned LLama2-7b with logical rules.}}
\end{table*}

\begin{table*}[hbt!]\footnotesize
\centering
\begin{tabular}{l|ccccccc}
\hline
\textbf{Models} &  \textbf{Medical Entity F1\% } &  \textbf{BLEU} &  \textbf{ROUGE-L } & \textbf{METEOR} & \textbf{Embedding Average} & \textbf{A-LEN} \\
\hline
GPT-2  & 7.88 & 0.0089 &  0.0604 & 0.0987 & 0.688 & 17.32   \\
BART  & 22.69 & 0.1154 &  0.1521 & 0.1723 & 0.729 & 29.19   \\
BioGPT  & 10.06 & 0.0241 &  0.0988 & 0.1212 & 0.701 & 21.25   \\
BioMistral  & 18.65 & 0.1358 &  0.1652 & 0.1798 & 0.735 & 54.50   \\
BioMedGPT-LM  & 14.25 & 0.1019 &  0.1427 & 0.1689 & 0.715 & 36.80   \\
\text{LLama2-Rule}   & 25.88 & 0.1558 &  0.1709 & 0.1864 & 0.741 & 66.23   \\
\text{LLama2+Rule}   & 27.12 & 0.2066 & 0.2047 & 0.1978 & 0.761 & 94.47   \\
\hline
\textsc{\textbf{MedLogic-AQA}} & \textbf{31.87} & \textbf{0.2284} &  \textbf{0.2513} & \textbf{0.1969} & \textbf{0.788} & {56.47}   \\
\hline
\end{tabular}
\caption{\label{tab:auto_results_mashqa}
{Automatic evaluation results on MASHQA dataset}}
\end{table*}

\begin{table*}[hbt!]\footnotesize
\centering
\begin{tabular}{l|ccccccc}
\hline
\textbf{Models} & \textbf{Fluency}  & \textbf{Adequacy} &  \textbf{Logical-Reasoning} & \textbf{Contextual-Consistency} \\ 
\hline
GPT-2 \cite{radford2019language} & 2.65 & 1.80 & 0.59 & 2.21 \\
BART \cite{lewis2020bart} & 3.21 & 2.18 & 1.92 & 2.81  \\
BioGPT \cite{luo2022biogpt} & 2.75 & 1.88 & 1.02 & 2.47  \\
BioMistral \cite{labrak2024biomistral} & 3.11 & 2.10 & 1.80 & 2.95  \\
BioMedGPT-LM \cite{luo2023biomedgpt} &  3.37 & 2.08 & 0.92 & 2.02  \\
\text{LLama2-Rule} \cite{touvron2023llama}  &  3.55   &  2.81 & 3.10 & 3.25 \\
\text{LLama2+Rule} \cite{touvron2023llama} &  4.10  &  3.11 & 3.75 & 3.65 \\
\hline
\textsc{\textbf{MedLogic-AQA}}  &  \textbf{4.41}  & \textbf{3.84} & \textbf{4.39} & \textbf{4.14} \\
\hline
\end{tabular}
\caption{\label{tab:human_results}
Human assessment results for the baseline and proposed model. The bolded values represent the best value.}
\end{table*}

\section{Results and Analysis}
In assessing the \textsc{MedLogic-AQA} performance, we employ both quantitative (Tables \ref{tab:auto_results_bioasq} and \ref{tab:auto_results_mashqa}) and qualitative analyses (Table \ref{tab:human_results}) to gauge its effectiveness in addressing medical queries.

\subsection{Automatic Evaluation}
Table \ref{tab:auto_results_bioasq} showcases the results of our automatic evaluation metrics on the \textit{BioASQ} dataset.

\textsc{MedLogic-AQA} demonstrates exceptional proficiency across various metrics, underscoring its adeptness in abstractive QA for biomedical queries. Notably, \textsc{MedLogic-AQA} achieves the highest scores for Medical Entity F1\% (38.47\%) and all BLEU levels, indicating precision in identifying medical entities and generating contextually relevant responses. Additionally, the model performs impressively in ROUGE-L, showcasing its ability to produce summaries closely aligned with reference summaries. The superior performance in embedding-based metrics, particularly in Embedding Average, underscores the model's effectiveness in generating meaningful contextual embeddings.

Similarly, Table \ref{tab:auto_results_mashqa} presents the outcomes of automatic evaluation metrics on the \textit{MASHQA} dataset. Here, \textsc{MedLogic-AQA} demonstrates consistent excellence across various metrics, showcasing its proficiency in abstractive QA for medical queries. An insightful observation from both the datasets reveals the consistent outperformance of \textsc{MedLogic-AQA} over baseline models across all the evaluation metrics. Particularly notable are the significant improvements in medical entity identification, summarization quality, and overall contextual understanding compared to the baseline models.

\subsection{Human Evaluation}

Table \ref{tab:human_results} presents the results of human evaluation, comparing baseline models with our proposed approach. 
In this assessment, our proposed models consistently outperforms the baseline models across various criteria, including Fluency, Adequacy, Logical-Reasoning and Context-consistency. 

The proposed model secures the highest ratings in Fluency (4.41), Adequacy (3.84), Logical-Reasoning (4.39), and Context-consistency (4.14), aligning with its superior performance in automatic evaluation metrics. These findings collectively affirm the effectiveness of the proposed model in generating contextually coherent, adequately informative, and logically sound responses to biomedical questions, as validated by both automatic and human evaluation.

\subsection{Result Analysis: Comparison of Answer Generation}

While analyzing the results obtained from both automatic evaluation metrics and human assessments, a clear distinction emerges between the performance of MedLogic-AQA and the baseline models. MedLogic-AQA consistently outperforms the baseline models across various evaluation metrics. For instance, consider the comparison between the answers generated by MedLogic-AQA and the baseline model "LLama + Rule" for the question "Is Apremilast effective for Behcet’s syndrome?", and ground truth answer was "Yes.  Apremilast was proven to be effective for treatment of Behcet’s syndrome."

\textbf{Answer generated by MedLogic-AQA}: "Yes, Apremilast is effective for Behcet’s syndrome. Apremilast is an immunomodulatory agent that works through phosphodiesterase 4 inhibition. A randomized controlled trial has shown that it is effective for the management of oral and genital ulcers and is generally well tolerated."

\textbf{Answer generated by "LLama + Rule" model}: "Apremilast, an oral small molecule inhibitor of phosphodiesterase 4 (PDE4), is in development for chronic inflammatory disorders, and has shown efficacy in psoriasis, psoriatic arthropathies, and Behçet's syndrome. Apremilast (Otezla(®)), an oral small molecule inhibitor of type-4 cyclic nucleotide phosphodiesterase (PDE-4), is under development with Celgene Corporation for the treatment of psoriatic arthritis, psoriasis, ankylosing spondylitis, Behçet's syndrome, atopic dermatitis, and rheumatoid arthritis."

While both the models provide responses, \textsc{MedLogic-AQA} generates a more detailed and logically coherent answer. It explains that Apremilast, as an immunomodulatory agent working through phosphodiesterase 4 inhibition, has shown efficacy in managing oral and genital ulcers based on a randomized controlled trial. This demonstrates the model's logical understanding capacity and its ability to reason during answer generation. In contrast, the baseline model's response lacks detailed explanation and logical inference regarding medical entities.

Result analysis of the LU model are shown in the Appendix \ref{sec:lu_result}.

\subsection{Case Study}
Evaluation figures mentioned in the tables underscore the effectiveness of our proposed technique, demonstrating its prowess in offering a more holistic comprehension of evidence and context. This section illuminates specific examples, with Table \ref{tab:case_study} showcasing two instances. The initial case study delves into the evaluation of Apremilast's efficacy for Behçet's syndrome. The responses from \textit{BART} and \textit{GPT2} succinctly confirm Apremilast's effectiveness, while \textit{Llama2-Rule} straightforwardly affirms this conclusion. In contrast, \textit{Llama2+Rule} delivers a more detailed response, highlighting Apremilast's role as an oral small molecule inhibitor of phosphodiesterase 4, currently in development for various inflammatory disorders, including Behçet's syndrome. The \textit{MedLogic-AQA} model echoes this sentiment, referencing a randomized controlled trial that validates Apremilast's efficacy in managing oral and genital ulcers associated with Behçet's syndrome. The inclusion of logical reasoning enriches the affirmation, aligning it with the broader context of Apremilast's mechanism of action and its potential applications in chronic inflammatory conditions.

Similarly, the second case study explores the diagnosis of Meigs' syndrome, characterized by a benign ovarian tumor accompanied by ascites and pleural effusion. For the models \textit{BART}, \textit{GPT2}, and \textit{Llama2-Rule}, concise affirmations underscore the consideration of Meigs' syndrome in the presence of specific symptoms. However, \textit{Llama2+Rule} and \textit{MedLogic-AQA} contribute more nuanced insights, elucidating the benign nature of Meigs' syndrome and underscoring the potential for misdiagnosis in cases with elevated CA-125 levels. These responses align with the broader medical context, showcasing a deeper understanding of Meigs' syndrome. The detailed context provided by \textit{Llama2\_finetune with rule} and \textit{MedLogic-AQA} enhances the overall comprehension of Meigs' syndrome, presenting a more comprehensive perspective on its diagnosis and potential pitfalls in clinical assessments.

\subsection{Error Analysis}
To provide a comprehensive investigation into the performance of the system, aiming to identify and understand the nature of errors encountered during evaluation, qualitative and quantitative error analysis is also performed.

\subsubsection{Qualitative Analysis}
In the qualitative analysis of errors, we delve into the specific instances where \textsc{MedLogic-AQA} produced incorrect or irrelevant answers.

\paragraph{Triples Generated by the LU Model}
\paragraph{LU Model-Generated Triples}
Table \ref{tab:triple_gen} illustrates the LU model's output for a question from the MASHQA dataset: "How do doctors diagnose delusional disorder?" Utilizing the 'Rule of Co-occurrence,' the model produces KG triples, such as ("delusional disorder", "affects", "psychotic disorders") and ("delusional disorder", "affects", "dopamine"). However, these triples do not directly address the query or provide relevant medical insights regarding diagnosis or treatment methods.

The presence of irrelevant triples highlights a constraint in the LU model's capacity to discern contextually significant associations and generate triples that align with the semantic context of the questions. Consequently, \textsc{MedLogic-AQA} may encounter difficulties in effectively utilizing the LU model's output to furnish coherent and informative answers to medical inquiries. Addressing this issue is crucial for enhancing the model's ability to provide accurate and relevant responses in medical question-answering tasks.

\paragraph{Impact of Incomplete Knowledge Graph}
Incomplete knowledge graph triples within datasets can significantly affect the performance of models. 

For example, in the BioASQ dataset, consider the question: "How does trimetazidine affect intracellular kinase signaling in the heart?" The generated triples from UMLS, such as ("injury", "affects", "function") and ("trimetazidine", "diagnoses", "mitogen"), indicate a lack of relevant information regarding trimetazidine's effects on intracellular kinase signaling in the heart. This deficiency in the knowledge graph triples may lead to inaccurate or incomplete responses from MedLogic-AQA, highlighting the importance of comprehensive and accurate knowledge representation in biomedical question answering.

\subsubsection{Quantitative Analysis}
In the quantitative analysis of MedLogic-AQA, two notable issues emerge. Firstly, semantic inadequacies within the ground truth responses pose challenges for evaluation. For instance, in response to the query regarding treatment options for osteoporosis spine fractures, the provided ground truth focuses on hip fractures instead, indicating a lack of alignment between the questions and provided answers. Secondly, instances of ambiguous answers arise, as seen in the response to the inquiry about tests for diagnosing hypertensive heart disease. While the ground truth offers a general overview, MedLogic-AQA's response delves into specific tests and treatment options, potentially introducing ambiguity due to variations in medical practice. Addressing these issues is crucial to enhance the accuracy and reliability of MedLogic-AQA in providing contextually relevant responses to medical queries.

\section{Conclusion}
Our work presents \textsc{MedLogic-AQA}, an innovative AQA system that addresses the inherent challenges of capturing complex logical structures within contextual information. By leveraging first-order logic-based rules and adopting a graph-based representation, our system demonstrates a significant advancement in enhancing reasoning capabilities for nuanced and intricate queries. It excels in generating contextually rich answers, surpassing the limitations of existing methods, hence, facilitating a more structured understanding of information by incorporating logical rules derived from the context. This ensures that the generated answers not only align with the given query but also adhere to logical constraints within the text.

As we move forward, continued research and refinement of MedLogic-AQA hold the potential to further elevate its performance and broaden its applicability across diverse domains, establishing a foundation for the next generation of advanced abstractive question answering system.

\section*{Limitations}
While MedLogic-AQA demonstrates promising performance in biomedical QA, several limitations should be acknowledged. Firstly, the model's reliance on pre-existing knowledge graphs and databases may result in limitations due to incomplete or outdated information, leading to inaccuracies in generated responses. Additionally, the model's performance may be constrained by the quality and coverage of the underlying knowledge sources. Secondly, the abstractive nature of \textsc{MedLogic-AQA} may occasionally lead to the generation of responses that deviate from the input query or lack specificity, particularly in complex medical scenarios requiring precise and detailed explanations. Furthermore, the model's performance may vary across different medical domains and specialties, depending on the availability and relevance of training data. Lastly, while efforts have been made to address biases in training data and model outputs, inherent biases in the underlying datasets and knowledge sources may still persist, potentially influencing the generated responses. For more information refer to the Appendix \ref{apx:limitation}.

\subsection*{Ethics Statement}
Our research adheres to ethical principles and guidelines to ensure responsible use of AI technologies in healthcare. We prioritize patient privacy, and confidentiality in data collection and usage. Furthermore, we strive to mitigate biases in our models and outputs by employing diverse and representative datasets, conducting rigorous evaluations, and transparently reporting limitations and uncertainties. Our goal is to develop AI-driven tools like \textsc{MedLogic-AQA} to augment, rather than replace, human expertise in medical decision-making, with a focus on improving patient outcomes and advancing medical research. We are committed to ongoing monitoring and evaluation of our models' impact to ensure ethical and responsible deployment in clinical settings.

\section{Acknowledgement}
Authors gratefully acknowledge the generous support for the project ``Percuro-A Holistic Solution for Text Mining``, sponsored by Wipro Ltd.

\begin{table*}[ht!]
\centering
\begin{tabular}{|p{0.15\linewidth}|p{0.85\linewidth}|}
\hline
\textbf{Context 1} &  Apremilast is an immunomodulatory agent that works through phosphodiesterase 4 inhibition.  A randomized controlled trial has shown that it is effective for the management of oral and genital ulcers and is generally well tolerated.  AREAS COVERED: This review provides a digest of all current experience and evidence about pharmacological agents recently described as having a role in the treatment of BS, including...(truncated)  \\
\hline
\textbf{Question} & Is \colorbox{pink}{\textit{\color{blue}\textbf{Apremilast}}} effective for \colorbox{pink}{\textit{\color{blue}\textbf{Behcet’s syndrome?}}}   \\
\textbf{Ideal Answer} & Yes.  \colorbox{pink}{\textit{\color{blue}\textbf{Apremilast}}} was proven to be effective for treatment of \colorbox{pink}{\textit{\color{blue}\textbf{Behcet’s syndrome.}}}   \\
\hline
\textbf{BART} & Yes \colorbox{pink}{\textit{\color{blue}\textbf{Apremilast}}} is effective for \colorbox{pink}{\textit{\color{blue}\textbf{Behcet’s syndrome}}}	\\
\hline
\textbf{GPT2} &  Yes it is effective for \colorbox{pink}{\textit{\color{blue}\textbf{Behcet’s syndrome}}} \\
\hline
\textbf{Llama2-Rule} & Yes, \colorbox{pink}{\textit{\color{blue}\textbf{Apremilast}}} is effective for \colorbox{pink}{\textit{\color{blue}\textbf{Behcet’s syndrome.}}} \\
\hline
\textbf{Llama2+Rule} & \colorbox{pink}{\textit{\color{blue}\textbf{Apremilast}}}, an oral small molecule inhibitor of \colorbox{pink}{\textit{\color{blue}\textbf{phosphodiesterase 4 (PDE4)}}}, is in development for \colorbox{pink}{\textit{\color{blue}\textbf{chronic inflammatory disorders}}}, and has shown efficacy in \colorbox{pink}{\textit{\color{blue}\textbf{psoriasis, psoriatic arthropathies, and Behçet's syndrome.}}}  Apremilast (Otezla(®)), an oral small molecule inhibitor of \colorbox{pink}{\textit{\color{blue}\textbf{type-4 cyclic nucleotide phosphodiesterase (PDE-4)}}}, is under development with Celgene Corporation for the treatment of \colorbox{pink}{\textit{\color{blue}\textbf{psoriatic arthritis, psoriasis, ankylosing spondylitis, Behçet's syndrome,}}} \colorbox{pink}{\textit{\color{blue}\textbf{atopic dermatitis, and rheumatoid arthritis.}}} \\
\hline
\textbf{MedLogic-AQA} &   Yes, \colorbox{pink}{\textit{\color{blue}\textbf{Apremilast}}} is effective for \colorbox{pink}{\textit{\color{blue}\textbf{Behcet’s syndrome.}}}  \colorbox{pink}{\textit{\color{blue}\textbf{Apremilast}}} is an immunomodulatory agent that works through \colorbox{pink}{\textit{\color{blue}\textbf{phosphodiesterase 4 inhibition.}}}  A randomized controlled trial has shown that it is effective for the management of \colorbox{pink}{\textit{\color{blue}\textbf{oral and genital ulcers}}} and is generally well tolerated. \\
\hline
\hline
\textbf{Context 2} &  Meigs' syndrome is a benign ovarian tumor associated with ascites and pleural effusion.  Although postmenopausal women with ovarian tumor, ascites, pleural effusion, and elevation of CA-125 levels probably have malignant ovarian tumors, Meigs' syndrome must be considered in the differential diagnosis....(truncated)  \\
\hline
\textbf{Question} & List symptoms of \colorbox{pink}{\textit{\color{blue}\textbf{Meigs' Syndrome.}}}   \\
\textbf{Ideal Answer} & \colorbox{pink}{\textit{\color{blue}\textbf{Meigs' syndrome}}} is a \colorbox{pink}{\textit{\color{blue}\textbf{benign ovarian tumor}}} associated with \colorbox{pink}{\textit{\color{blue}\textbf{ascites and pleural effusion.}}}   \\
\hline
\textbf{BART} & \colorbox{pink}{\textit{\color{blue}\textbf{Meigs' syndrome}}} should be considered at the differential diagnosis for a patient with \colorbox{pink}{\textit{\color{blue}\textbf{pelvic mass}}} \\
\hline
\textbf{GPT2} &  The \colorbox{pink}{\textit{\color{blue}\textbf{Meigs syndrome}}} must be evoked in case of presence of a typical triad \\
\hline
\text{Llama2-Rule} & \colorbox{pink}{\textit{\color{blue}\textbf{Meigs' syndrome}}} is a \colorbox{pink}{\textit{\color{blue}\textbf{benign ovarian tumor}}} associated with \colorbox{pink}{\textit{\color{blue}\textbf{ascites and pleural effusion}}}. \\
\hline
\text{Llama2+Rule} & \colorbox{pink}{\textit{\color{blue}\textbf{Meigs' syndrome}}} is a \colorbox{pink}{\textit{\color{blue}\textbf{benign ovarian tumor}}} associated with \colorbox{pink}{\textit{\color{blue}\textbf{ascites and pleural effusion}}}.  Although \colorbox{pink}{\textit{\color{blue}\textbf{postmenopausal women}}} with \colorbox{pink}{\textit{\color{blue}\textbf{ovarian tumor, ascites, pleural effusion, and elevation of CA-125}}} levels probably have malignant ovarian tumors, Meigs' syndrome must be considered in the differential diagnosis. \\
\hline
\textbf{MedLogic-AQA} &   \colorbox{pink}{\textit{\color{blue}\textbf{Meigs' syndrome}}} is a \colorbox{pink}{\textit{\color{blue}\textbf{benign ovarian tumor}}} associated with \colorbox{pink}{\textit{\color{blue}\textbf{ascites and pleural effusion}}}.  Although \colorbox{pink}{\textit{\color{blue}\textbf{postmenopausal women}}} with \colorbox{pink}{\textit{\color{blue}\textbf{ovarian tumor, ascites, pleural effusion, and elevation of CA-125}}} levels probably have \colorbox{pink}{\textit{\color{blue}\textbf{malignant ovarian tumors}}}, Meigs' syndrome must be considered in the differential diagnosis.  Meigs' syndrome should be considered at the differential diagnosis for a patient with \colorbox{pink}{\textit{\color{blue}\textbf{pelvic mass, pleural effusion and ascites with normal cytology}}}, increased CA125 levels.  When benign ovarian fibroma is associated with ascites and/or pleural effusion it is termed Meigs syndrome. \\
\hline
\end{tabular}
\caption{
\label{tab:case_study} Case Study: Pink box represents medical entities found in sentences}
\end{table*}

\onecolumn
\appendix
\newpage
\section{Dataset}\label{apx:dataset}
\subsection{BioASQ Dataset}
The BioASQ Task 10b Phase B (QA task) dataset \cite{nentidis2022overview}, meticulously designed for biomedical question-answering. It centers around two primary tasks: biomedical semantic indexing (Task A) and question-answering (Task B). Our focus lies on the dataset specifically tailored for the QA task. The BioASQ Task 10b Phase B dataset consists of biomedical questions and relevant snippets. Participants in the challenge must provide either the exact answer or the ideal answer based on the given snippets. The questions and answers in the dataset are carefully constructed by a team of biomedical experts from across Europe, ensuring their quality and relevance.
The questions in the dataset are categorized into four groups: yes/no questions, factoids, lists, and summaries. Participants are expected to provide the ideal answer for each question, depending on its specific category and requirements.
The content of BioASQ task 10b Phase B dataset are:
\begin{itemize}
    \item \textbf{Questions:} The dataset includes a diverse set of biomedical questions. These questions can be categorized into four main types:
\begin{itemize}
    \item \textbf{Yes/No Questions:} Questions that require a binary "yes" or "no" answer.
    \item \textbf{Factoids:} Questions that seek specific factual information often require a concise answer.
    \item \textbf{Lists:} Questions that request a list of items, such as medications or diseases.
    \item \textbf{Summaries:} Questions that ask for a summary or synthesis of information.
\end{itemize}
    \item \textbf{Snippets:} For each question, the dataset provides relevant text snippets from biomedical sources. These snippets serve as the context from which answers should be derived. These snippets are used as context in our case.
    \item \textbf{Answers:} The dataset contains both "exact" answers and "ideal" answers. "Exact" answers are the precise answers to the questions, while "ideal" answers represent the most informative and relevant responses based on the context. We use the ideal answers as our ground truth answers.
\end{itemize}
The detailed dataset statistics are given in Table \ref{tab:data_stats_bioasq}.
\begin{table}[hbt!]\footnotesize
\centering
\begin{tabular}{|c|c|}
\hline
\textbf{Background Data} & \textbf{Unique Value} \\ \hline
QA pairs                 & 4,232                 \\ \hline
Average Question token   & 9                   \\ \hline
Average Answer token     & 37                  \\ \hline
Average context token       & 342                    \\ \hline
\end{tabular}
\caption{Detailed statistics of BioASQ Dataset}
\label{tab:data_stats_bioasq}
\end{table}

\subsection{MASH-QA Dataset}
The MASH-QA dataset consists of consumer healthcare questions gathered from the well-known health website WebMD. The website includes content from a wide range of consumer healthcare sectors. The website’s healthcare sections offer questions regarding frequent healthcare difficulties that people confront. It is the largest available dataset having around 25K question-answer pairs in the medical domain. The detailed dataset statistics are given in Table \ref{tab:data_stats_mashqa}.

\begin{table}[hbt!]\footnotesize
\centering
\begin{tabular}{|c|c|}
\hline
\textbf{Background Data} & \textbf{Unique Value} \\ \hline
QA pairs                 & 25,289                 \\ \hline
Average Question token   &  25.8                  \\ \hline
Average Answer token     & 67.2                  \\ \hline
Average context token       & 696.2                    \\ \hline
\end{tabular}
\caption{Detailed statistics of MASHQA Dataset}
\label{tab:data_stats_mashqa}
\end{table}

\subsection{Dataset Preparation}
We prepared our datasets through two key processes: converting the MASHQA dataset into an Abstractive QA format and pre-processing the dataset for the Logic Understanding (LU) module.

\paragraph{Converting MASHQA Dataset into Abstractive QA Dataset:}
To transform the MASHQA dataset into an abstractive form, we utilized the "chatgpt\_paraphraser\_on\_T5\_base" model\footnote{\href{https://huggingface.co/humarin/chatgpt_paraphraser_on_T5_base}{humarin/chatgpt\_paraphraser\_on\_T5\_base}}, which is built upon the T5-base architecture. This model employs transfer learning to generate paraphrases and leverages ChatGPT for the conversion process. 

For instance, given the question "What is hypertensive heart disease?" and its extracted answer "It refers to a group of disorders that includes heart failure, ischemic heart disease, and left ventricular hypertrophy," the abstractive answer becomes "Heart failure, ischemic heart disease, and left ventricular hypertrophy are among the disorders that fall under this category."

\paragraph{Dataset Pre-processing for LU Module:}
For the Logic Understanding (LU) module, we employed the Unified Medical Language System (UMLS)\cite{bodenreider2004unified} to extract knowledge graph triplets. These triplets were then subjected to logical rules to filter and create new triplets based on predefined logical rules.

For instance, for the question "Which disease can be treated with Relugolix," the UMLS generated triples include (androgen deprivation therapy, affects, uterine fibroids), (androgen deprivation therapy, treats, pain), (androgen deprivation therapy, prevents, endometriosis), among others. These triples were further processed by logical rules, resulting in
\textbf{Rule of Co-occurrence:} (androgen deprivation therapy, affects, disorders), (androgen deprivation therapy, affects, endometriosis), (androgen deprivation therapy, affects, suppression),...(truncated) 
\textbf{Rule of Prevention and Causation:} (androgen deprivation therapy, prevents, endometriosis), (androgen deprivation therapy, prevents, disorders), (external beam radiotherapy, prevents, endometriosis),...(truncated)
\textbf{Rule of Treatment and Classification:} (androgen deprivation therapy, treats, pain), (androgen deprivation therapy, treats, endometriosis), (androgen deprivation therapy, treats, prostate cancer)... (truncated)

\subsection{Prompt used to fine-tune LU module}
To fine-tune the Llama2-7b model, we utilize a prompt consisting of specific rules and context.
The prompt includes rules, such as co-occurrence, prevention and causation, treatment and classification, diagnosis and interaction, conjunction, and disjunction. This fine-tuning process enables the model to generate knowledge graph triples to support answers for given questions and contexts effectively.

\section{Baseline Models: Detailed Descriptions and Comparisons}
\label{apx:baseline-details}

In this section, we provide detailed descriptions of the baseline models used in our experiments, including their training methodologies and how they differ from our proposed \textsc{MedLogic-AQA} system.

\paragraph{BART \cite{lewis2020bart}} BART is a sequence-to-sequence model pre-trained as a denoising autoencoder. It is fine-tuned on the question-answering task using the input format $x = [q_i + c_i]$, where $q_i$ is the question and $c_i$ is the context. The model is then trained to generate the answer $y = a_i$. This baseline does not incorporate explicit logical rules, making it purely a text-based QA system.

\paragraph{GPT2 \cite{radford2019language}} GPT-2 is a transformer-based language model trained on a large corpus of general-domain text. Similar to BART, it is fine-tuned for the question-answering task using the input $x = [q_i + c_i]$ and output $y = a_i$. GPT-2 lacks any domain-specific medical knowledge and logical reasoning capabilities, focusing solely on context-based responses.

\paragraph{BioGPT \cite{luo2022biogpt}} BioGPT is a variant of GPT-2 specifically pre-trained on biomedical text. It is designed to understand and generate domain-specific language. We fine-tune BioGPT on the input format $x = [q_i + c_i]$, using the output $y = a_i$. While it possesses a deeper understanding of medical terminology compared to GPT-2, it still does not incorporate structured logical rules.

\paragraph{BioMistral-7B \cite{labrak2024biomistral}} BioMistral-7B is a recent large language model (LLM) pre-trained on biomedical data. We adapt it to the question-answering task using $x = [q_i + c_i]$ and $y = a_i$. This model leverages a large-scale biomedical corpus for enhanced comprehension but does not include explicit logical reasoning.

\paragraph{BioMedGPT-LM-7B \cite{luo2023biomedgpt}} BioMedGPT-LM-7B is a language model pre-trained on a diverse range of biomedical sources, including academic papers and medical guidelines. Similar to other baselines, it is fine-tuned for question-answering using $x = [q_i + c_i]$ and $y = a_i$. While it is more specialized in the biomedical domain, it does not explicitly model logical relationships.

\paragraph{LLama2-Rule} This baseline involves fine-tuning the LLama2 model \cite{touvron2023llama} using the standard input $x = [q_i + c_i]$ and output $y = a_i$. This setup serves as a control baseline that does not incorporate logical rules. The focus is on evaluating LLama2's performance when fine-tuned with medical text alone.

\paragraph{LLama2+Rule} This variant of LLama2 is fine-tuned using logical rules as additional input, i.e., $x = [q_i + c_i + R_i]$, where $R_i$ represents the set of six predefined logical rules. The output remains $y = a_i$. By including the logical rules, this baseline aims to understand the impact of structured logic on the question-answering performance.

\paragraph{Comparison to \textsc{MedLogic-AQA}} Unlike these baselines, our proposed \textsc{MedLogic-AQA} system follows a two-stage training process. In the first stage, the Logic Understanding (LU) model is fine-tuned to generate logical triples ($y = lt$) from the input $x = [q_i + c_i + a_i + R_i]$, which helps the model learn structured logical representations. In the second stage, this structured logical information is leveraged by the Answer Quality Assurance (AQA) model, which uses the input $x = [q_i + c_i + R_i]$ to generate logically coherent answers $y = a_i$. This two-stage approach allows \textsc{MedLogic-AQA} to produce answers that are not only contextually relevant but also logically consistent, differentiating it from baselines that do not incorporate explicit logical reasoning.
\section{Human Evaluation}\label{apx:human_eval}
To assess the quality of answers generated by our model, we carefully selected 120 generated answers along with their corresponding questions, contexts, and ground-truth answers from the BioASQ dataset. For the human evaluation, we recruited a panel of five evaluators with diverse backgrounds. Three of the evaluators hold post-graduate qualifications in linguistics and possess significant experience in tasks related to natural language generation. Additionally, two evaluators have MD degrees, providing domain-specific expertise. Evaluation were performed in two phases. In the first phase, three evaluators evaluate the answers generated by all baselines and \textsc{MedLogic-AQA} as per context, question and ground-truth answer. Then in the second phase, these evaluation were cross checked by two medical experts possessing MD degrees. Evaluations where deviating high or low scores were found were re-evaluated by these medical experts as per their own domain specific knowledge, given context, question and answer. By including evaluators with both linguistic and medical backgrounds, we ensured a comprehensive assessment of answer quality. Our evaluators were not sourced from Mechanical Turk but were specifically recruited based on their qualifications and expertise.

\section{Derivation of Logical Rules}\label{apx:logic_rule}

The process of deriving the six logical rules involved several steps to ensure their effectiveness and applicability across diverse medical knowledge domains. Initially, we constructed a medical KG using the UMLS, which contains a comprehensive set of semantic relations between medical entities. Out of the 54 available semantic relations in UMLS, we carefully selected seven key relationships, including "co-occurs-with," "prevent," "treat," "diagnosis," "interacts-with," "affects," and "causes."

This selection was made to strike a balance between computational efficiency and relevance, as processing a larger number of relations would require significant computing resources and may introduce irrelevant or redundant information into the KG. Subsequently, we created the KG based on this curated set of relationships, using contextual information extracted from medical documents.

For instance, when processing a context related to the treatment of uterine fibroids with Relugolix, the KG received contained triples such as ["androgen\_deprivation\_therapy", "affects", "uterine\_fibroids"] and ["heavy\_menstrual\_bleeding", "co-occurs-with", "uterine\_fibroids"], among others. 
After thorough analysis of these KG triples, we identified patterns indicating logical relationships between entities, such as the rule "co-occurs-with(X, Y) $\land$ affects(Y, Z) leads to affects(X, Z)." These rules underwent verification by medical domain specialists to ensure their accuracy and relevance. Ultimately, we finalized six rules out of the initial twelve, as the rejected six rules did not yield any logical triples.

\section{Two-Stage Model Training: LU and AQA Models}
\label{luaqa}
The Logic Understanding (LU) model and the Answer Quality Assurance (AQA) model, although stemming from the same base architecture, serve different purposes and thus operate with different inputs and outputs during their respective training stages.

\paragraph{Stage 1 (LU Model Training)} In this stage, the LU model is fine-tuned using the following input and output format:

\textbf{Input:} $x = [q_i + c_i + a_i + R_i]$, where $q_i$ is the question, $c_i$ is the context, $a_i$ is the answer, and $R_i$ represents a set of six predefined logical rules. \\
\textbf{Output:} $y = lt$, denoting the logical triples.

This stage focuses on enabling the model to learn logic graphs that capture the logical relationships within the input data. The probability distribution, $p_{\theta}(y|x)$, approximates the likelihood of generating logical triples given the input sequence.

\paragraph{Stage 2 (AQA Model Training)} The same base model is further fine-tuned to become the AQA model with a different input and output format:

\textbf{Input:} $x = [q_i + c_i + R_i]$. \\
\textbf{Output:} $y = a_i$.

The first-stage model (LU model) enhances the second-stage training (AQA model) by providing structured logical information that helps the model understand the relationships and dependencies in the data. This structured representation aids the AQA model in generating answers that are not only contextually relevant but also logically coherent.

\paragraph{Training Rationale and Relation to Previous Work} The two-stage training strategy is inspired by the approach used in pre-trained language models (LLMs), which first learn general interactions between words and then leverage this knowledge when fine-tuned on downstream tasks. Specifically, this approach is motivated by the observations in the Phi-1.5 paper \cite{li2023textbooks}, which demonstrated that a double-fine-tuned model (trained on specialized fine-grained data) could effectively utilize the learning from a single-fine-tuned model (trained on raw data) to call the correct libraries when generating code. 

By training the LU model first, we ensure that the parameters learn to perform logical reasoning. These parameters are then leveraged in the second stage to develop the AQA model, ensuring that the final model can generate logically coherent answers.

\section{Limitations of Models and Analysis}\label{apx:limitation}

The evaluation of our model's performance revealed several limitations and areas for improvement. 

Firstly, an error analysis conducted on the BioASQ dataset, as shown in Table \ref{tab:factual_error}, highlighted a factual knowledge problem. Despite the ideal answer indicating that splicing speckles contain little detectable transcriptional activity, some models, including \textsc{MedLogic-AQA}, initially asserted that splicing speckles are not associated with transcription. However, upon further examination of the context, these models provided detailed explanations indicating the presence of transcription-related processes within splicing speckles. This discrepancy underscores a potential limitation in the models' ability to accurately infer factual information solely based on the provided context, necessitating the integration of logical reasoning to refine and rectify such errors.

Furthermore, an error analysis from the MASHQA dataset, presented in Table \ref{tab:analysis_mashqa_1}, revealed a bias in our model's responses towards infants. While the ideal answer underscored the risk of dehydration in both adults and young children, specifically infants, the \textsc{MedLogic-AQA} model's responses focused solely on infants, neglecting to provide guidance for adults or other age groups. This bias towards infants could potentially lead to inadequate or incomplete information for caregivers and adults facing similar situations.

Additionally, the deviation of answers, as seen in the example of Table \ref{tab:analysis_mashqa_2}, was evident. While the ideal answer addressed the long-term effects of chemotherapy on weight, emphasizing the challenges faced by individuals, particularly those undergoing breast cancer treatment, the \textsc{MedLogic-AQA} responses diverged by primarily focusing on the effects of chemotherapy on hair and the risk of permanent baldness. This deviation from the intended scope of the question suggests a slight hallucination or misinterpretation of the context, highlighting the need for improved model robustness and comprehension of nuanced medical queries.

Upon further analysis, differences in reasoning abilities were observed between the BioASQ and MASHQA datasets. The BioASQ dataset demonstrated higher fluency and logical explanation in its responses compared to the MASHQA dataset, prompting further investigation into the underlying factors.

These differences in reasoning abilities can be attributed to variations in the performance of the Logic Understanding (LU) model, particularly in the quality of the knowledge graph (KG) triplets generated by the Unified Medical Language System (UMLS). The LU model exhibited better performance in the BioASQ dataset, generating more accurate and relevant triplets conducive to logical reasoning. Conversely, the MASHQA dataset showed slightly lower performance, likely due to limitations or inconsistencies in the KG triplets generated by the UMLS. These disparities may have affected the LU model's ability to infer logical relationships effectively, resulting in less coherent and contextually relevant responses.

\begin{table*}[ht!]
\centering
\begin{tabular}{|p{0.15\linewidth}|p{0.85\linewidth}|}
\hline
\textbf{Context} &  We show here that RNA splicing speckled domains (splicing speckles) fluctuate in constrained nuclear volumes and remodel their shapes. We present a model where recycling splicing factors return as part of small sub-speckles from distal sites of RNA processing to larger splicing speckles by a directed ATP-driven mechanism through interchromatin spaces. Analysis of a HeLa cell line stably expressing EYFP-NHPX showed that the nucleolar accumulation of NHPX was preceded by its transient accumulation in splicing speckles....(truncated)  \\
\hline
\textbf{Question} & Are splicing speckles associated with transcription?   \\
\textbf{Ideal Answer} & Speckles contain little detectable transcriptional activity.   \\
\hline
\textbf{BART} & plicing speckles are major nuclear domains rich in components of the splicing machinery and polyA(+) RNA.\\
\hline
\textbf{GPT2} &  yes, an inducible splicing speckle, R/H, serves as part of pre-mRNA metabolism in BPS \\
\hline
\textbf{Llama2-Rule} & Yes, splicing speckles are associated with transcription. \\
\hline
\textbf{Llama2+Rule} & Splicing speckles are major nuclear domains rich in components of the splicing machinery and polyA(+) RNA.  Although speckles contain little detectable transcriptional activity, they are found preferentially associated with specific mRNA-coding genes and gene-rich R bands, and they accumulate some unspliced pre-mRNAs RNA polymerase II transcribes mRNAs and is required for splicing, with some reports suggesting that the inactive complexes are stored in splicing speckle In normal cell growth conditions GFPeIF4A-III was mainly nucleoplasmic, but in hypoxia stress conditions it moved to the nucleolus and splicing speckles.  Localization of eIF4A-III in the nucleolus and splicing speckles is an indicator of plant stress.   Using antibodies raised against mouse RBM6 to immunostain mammalian cell lines we found that the endogenous protein was both distributed diffusely in the nucleus and concentrated in a small number of nuclear foci that corresponded to splicing speckles/interchromatin granule clusters (IGCs Subnuclear targeting of the RNA-binding motif protein RBM6 to splicing speckles and nascent transcripts \\
\hline
\textbf{MedLogic-AQA} &   No, splicing speckles are not associated with transcription.  Splicing speckles are major nuclear domains rich in components of the splicing machinery and polyA(+) RNA.  Although speckles contain little detectable transcriptional activity, they are found preferentially associated with specific mRNA-coding genes and gene-rich R bands, and they accumulate some unspliced pre-mRNAs.  RNA polymerase II transcribes mRNAs and is required for splicing, with some reports suggesting that the inactive complexes are stored in splicing speckle.  Localization of eIF4A-III in the nucleolus and splicing speckles is an indicator of plant stress.  Analysis of a HeLa cell line stably expressing EYFP-NHPX showed that the nucleolar accumulation of NHPX was preceded by its transient accumulation in splicing speckles.  In vivo analysis of NHPX reveals a novel nucleolar localization pathway involving a transient accumulation in splicing speckles. \\
\hline
\end{tabular}
\caption{
\label{tab:factual_error}Error Analysis from BioASQ dataset indicating a factual knowledge problem. The proposed model initially provides a "No" answer, but later transitions to a "Yes" with logical reasoning.}
\end{table*}

\begin{table*}[ht!]
\centering
\begin{tabular}{|p{0.15\linewidth}|p{0.85\linewidth}|}
\hline
\textbf{Context} &  Nausea is an uneasiness of the stomach that often comes before vomiting. Vomiting is the forcible voluntary or involuntary emptying (\"throwing up\") of stomach contents through the mouth. Nausea and vomiting are not diseases, but they are symptoms of many conditions such as: Motion sickness or seasickness Early stages of pregnancy....(truncated)  \\
\hline
\textbf{Question} & How can I get dehydration from vomiting?   \\
\textbf{Ideal Answer} & The risk of dehydration is lower for adults as they can usually sense signs of thirst and dryness, but young children are more likely to become dehydrated, especially if they have diarrhea. Adults who care for sick children should be alert to dehydrating symptoms, such as dry lips and mouth, sunken eyes, and rapid breathing or pulse. In infants, it is important to monitor for decreased urine production and a sunkken fontanelle in newborns.   \\
\hline
\textbf{BART} & Take a child over age six years to the doctor if vomiting lasts one day, diarrhea combined with vomiting lasts for more than 24 hours, this are any signs of dehydration \\
\hline
\textbf{GPT2} &  consult a doctor if vomiting occurs for more than one day \\
\hline
\textbf{Llama2-Rule} & Adults are at risk of becoming dehydrated due to the presence of dry lips and mouth, sunken eyes, and rapid breathing or pulse. In infants, also watch for decreased urination and a sunken fontanelle (soft spot on top of the baby's head). Recurrent vomiting in pregnancy can lead to a serious condition called hyperemesis gravidarum in which the mother may develop fluid and mineral imbalances \\
\hline
\textbf{Llama2+Rule} & Adults should consult a doctor if vomiting occurs for more than one day, diarrhea and vomiting lasts for more than 24 hours, there are signs of dehydration, there is a fever higher than 101 degrees, or the child hasn't urinated for six hours. Take a child over age six years to the doctor if vomiting lasts one day, diarrhea combined with vomiting lasts for more than 24 hours, there are any signs of dehydration, there is a fever higher than 101 degrees, or the child hasn't urinated for six hours. \\
\hline
\textbf{MedLogic-AQA} &   Infants are more likely to experience dehydration due to recurrent vomiting during pregnancy, which can lead to life-threatening conditions such as hyperemesis gravidarum. \\
\hline
\end{tabular}
\caption{
\label{tab:analysis_mashqa_1}  Error Analysis from MASHQA dataset illustrating a bias towards infants in our model's response}
\end{table*}

\begin{table*}[ht!]
\centering
\begin{tabular}{|p{0.15\linewidth}|p{0.85\linewidth}|}
\hline
\textbf{Context} & Some side effects of chemotherapy only happen while you're having treatment and disappear quickly after it's over. But others can linger for months or years, or may never completely go away. Watch out for signs of chemo's long-term changes, and let your doctor know...(truncated)  \\
\hline
\textbf{Question} & How can chemotherapy affect your weight?   \\
\textbf{Ideal Answer} & Some chemo treatments can cause weight gain. However, it may not disappear even after the treatment is over. For example, some breast cancer medications result in loss of muscle mass and fat accumulation, making it more challenging to lose weight. Consult your doctor or nutritionist to help you shed some extra pounds by performing strength training exercises.   \\
\hline
\textbf{BART} & Some chemo treatments can cause weight gain. However, it may not disappear even after the treatment is over. \\
\hline
\textbf{GPT2} &  some drugs that treat breast cancer can cause you to lose muscle and gain fat, which makes it more difficult to lose weight. \\
\hline
\textbf{Llama2-Rule} & The weight may remain on your body even after treatment has ended. Some types of chemotherapy can cause you to lose muscle and gain fat, which makes it more difficult to lose weight. Incorporate strength training exercises into your workout routine to help you shed some pounds. Consult a dietitian or nutritionist for advice.\\
\hline
\textbf{Llama2+Rule} & The weight may remain on your body even after treatment, but it is possible that the effects of chemotherapy on your hair follicles will wear off over time. In rare cases, such as after years of strong chemotherapy, you may become permanently bald. Some types of chemo can cause you to lose muscle and gain fat, which makes it more difficult to lose weight. Work with a dietitian or nutritionist and add strength training exercises to your workout routine to help you shed some pounds. You may remember the fatigue that hit you while you were getting chemo. But some people still feel extremely tired after treatment ends. Even when you get enough rest, you may still feel sl \\
\hline
\textbf{MedLogic-AQA} & The weight of your hair may not return to its original state after treatment. It may be different in texture, shape, and color. Over time, as the effects of the chemo on your hair follicles wear off, your hair will likely return to its original state. In rare cases, such as after years of strong chemotherapy, your hair follicles can shut down, preventing new hair from growing and causing permanent baldness. Some types of chemo can cause you to put on extra pounds. You may find that the weight stays on even after treatment ends. For instance, some drugs that treat breast cancer can cause you to lose muscle and gain fat, which makes it more difficult to lose weight. Work with a dietitian or nutritionist and add strength training exercises to your workout routine to help you. \\
\hline
\end{tabular}
\caption{
\label{tab:analysis_mashqa_2}  Error Analysis from MASHQA dataset highlighting the answer deviation problem (which is slightly related to hallucination).}
\end{table*}

\section{Results of Logic Understanding Module}\label{sec:lu_result}
We have assessed the performance of the LU module using both the BioASQ and MASHQA datasets. Table \ref{tab:lu_results} presents the results of our automatic evaluation metrics.

From Table \ref{tab:lu_results}, it is evident that the LU model performs better on the BioASQ dataset across all the metrics compared to the MASHQA dataset. However, both the datasets present persistent challenges. Ambiguously defined answer spans and semantic inadequacies notably contribute to errors in MASHQA. In contrast, the BioASQ dataset benefits from meticulously crafted questions and answers by biomedical experts, resulting in more relevant knowledge triples. 

The LU model's performance underscores these challenges, with reduced coverage of domain-specific concepts and impaired inference capabilities, especially in tasks necessitating intricate reasoning within biomedical contexts. To mitigate these limitations and bolster the LU model's performance across a spectrum of biomedical QA tasks, enhancements in dataset annotation and model training strategies are imperative.

An illustrative example in Table \ref{tab:lu_analysis} highlights the module's efficiency in analyzing which logical rule should be applied and generating knowledge triples based on that rule. When provided with the answer generated by \textbf{MedLogicQA} model, "Yes, Apremilast is effective for Behcet’s syndrome. Apremilast is an immunomodulatory agent that works through phosphodiesterase 4 inhibition. A randomized controlled trial has shown that it is effective for the management of oral and genital ulcers and is generally well tolerated," the LU module generates knowledge triples based on the \textit{Rule of Diagnosis and Interaction} viz. are (apremilast, diagnoses, immunomodulatory agent), (apremilast, diagnoses, phosphodiesterase), (apremilast, diagnoses, treatment), (apremilast, diagnoses, phosphodiesterase 4), (apremilast, diagnoses, treatment agent), (apremilast, diagnoses, phosphodiesterase 4 inhibition). 

\begin{table}[ht]
\centering
\begin{tabular}{l|cccccc}
\hline
\textbf{Evaluation Metric} & \textbf{Bleu\_1} & \textbf{Bleu\_2} & \textbf{Bleu\_3} & \textbf{Bleu\_4} & \textbf{METEOR} & \textbf{ROUGE\_L} \\
\hline
\textbf{BioASQ} & 0.115723 & 0.071302 & 0.046654 & 0.028698 & 0.128755 & 0.099447 \\
\textbf{MASHQA} & 0.105723 & 0.069302 & 0.041654 & 0.024698 & 0.112755 & 0.08844 \\
\hline
\end{tabular}
\caption{LU model results on BioASQ and MASHQA datasets.}
\label{tab:lu_results}
\end{table}

\begin{table}[htbp]
\centering
\begin{tabular}{|p{0.25\linewidth}|p{0.7\linewidth}|}
\hline
\textbf{Module} & \textbf{Results} \\
\hline
\textbf{Question} & Is Apremilast effective for Behcet’s syndrome? \\
\hline
\textbf{MedLogic-AQA} & Yes, Apremilast is effective for Behcet’s syndrome. Apremilast is an immunomodulatory agent that works through phosphodiesterase 4 inhibition. A randomized controlled trial has shown that it is effective for the management of oral and genital ulcers and is generally well tolerated. \\
\hline
\textbf{LU Module} & Rule of Diagnosis and Interaction: [(apremilast, diagnoses, immunomodulatory agent), (apremilast, diagnoses, phosphodiesterase), (apremilast, diagnoses, treatment), (apremilast, diagnoses, phosphodiesterase 4), (apremilast, diagnoses, treatment agent), (apremilast, diagnoses, phosphodiesterase 4 inhibition)] \\
\hline
\textbf{Medical KG} & Rule of Co-occurrence: [(behcet, affects, adverse\_events), (oral\_ulcer, affects, adverse\_events)], Rule of Treatment and Classification: [(apremilast, treats, adverse\_events), (alemtuzumab, treats, adverse\_events), (tocilizumab, treats, adverse\_events), (ustekinumab, treats, adverse\_events)], Rule of Diagnosis and Interaction: [(tocilizumab, diagnoses, apremilast), (alemtuzumab, diagnoses, tocilizumab), (alemtuzumab, diagnoses, ustekinumab)] \\
\hline
\end{tabular}
\caption{LU model results on BioASQ and MASHQA datasets.}
\label{tab:lu_analysis}
\end{table}

\section{Hyperparameters}\label{sec:hyperparameters}
Table \ref{tab:hyperparameters} provides an organized representation of the hyperparameters used in the experiment, categorized by model setup and training parameters.

\begin{table}[htbp]
\centering
\caption{Hyperparameters Used in the Experiment}
\begin{tabular}{|l|l|}
\hline
\textbf{Model Setup}            &   \\ \hline
BitsAndBytesConfig              &   \\ 
load\_in\_4bit                  & True \\
bnb\_4bit\_quant\_type         & "nf4"  \\
bnb\_4bit\_compute\_dtype      & float16  \\
bnb\_4bit\_use\_double\_quant  & False   \\ \hline
\textbf{Training}               &       \\ \hline
LoRA Configuration              &       \\
lora\_alpha                    & 16      \\
lora\_dropout                  & 0.1    \\
lora\_r                        & 64     \\
task\_type                     & "CAUSAL\_LM"   \\ \hline
TrainingArguments               &    \\ \hline
per\_device\_train\_batch\_size & 8   \\
per\_device\_eval\_batch\_size  & 4  \\
gradient\_accumulation\_steps   & 2    \\
gradient\_checkpointing        & True   \\
optim                           & "paged\_adamw\_32bit" \\
logging\_steps                 & 25    \\
learning\_rate                 & 2e-4    \\
fp16                           & False    \\
bf16                           & False    \\
max\_grad\_norm                & 0.3     \\
num\_train\_epochs             & 15    \\
max\_steps                     & -1    \\
evaluation\_strategy           & "steps"  \\
eval\_steps                    & 0.2    \\
warmup\_ratio                  & 0.03        \\
weight\_decay                  & 0.001   \\
lr\_scheduler\_type            & "cosine"  \\
seed                            & 42        \\ \hline
\end{tabular}
\label{tab:hyperparameters}
\end{table}

\begin{table*}[ht!]
\centering
\begin{tabular}{|p{0.15\linewidth}|p{0.85\linewidth}|}
\hline
\textbf{Context 1} &  Delusional disorder, previously called paranoid disorder, is a type of serious mental illness called a psychotic disorder. People who have it can't tell what's real from what is imagined. Delusions are the main symptom of delusional disorder. They're unshakable beliefs in something...(truncated)  \\
\hline
\textbf{Question} & How do doctors diagnose delusional disorder?   \\
\textbf{Ideal Answer} & If you exhibit symptoms of delusional disorder, your doctor may conduct a medical examination and comprehensive medical history. Although there are no lab tests to diagnose delusion disorder in general, the doctor can sometimes use imaging studies or blood tests as diagnostic tools to help diagnose symptoms.   \\
\hline
\textbf{KG triple for Co-occurrence} & ['delusional disorder', 'affects', 'psychotic disorders'], ['delusional disorder', 'affects', 'mental illness'], ['delusional disorder', 'affects', 'dopamine']\\
\hline
\textbf{KG triple for Conjunction} & ['delusions', 'co-occurs\_with', 'psychological factors'], ['delusions', 'co-occurs\_with', 'perceptions'], ['delusions', 'co-occurs\_with', 'insight'], ['delusions', 'co-occurs\_with', 'hallucinations']\\
\hline
\end{tabular}
\caption{
\label{tab:triple_gen}Qualitative Analysis: LU Model's Failure to Generate Appropriate KG Triple.}
\end{table*}

\section{Experimental Design and Hypotheses}\label{apx:exp_design}

Our research aims to investigate the model's inherent reasoning capabilities and its adaptability to different inputs. We hypothesize that fine-tuning the LU model with diverse outputs enables it to effectively perform logical reasoning tasks. 
This study is motivated by the need to explore the model's ability to prioritize and weigh rules based on context. We posit that the fine-tuning process will facilitate the organization and consolidation of knowledge acquired during the training of the LU module. This knowledge, embedded within the parameters of the LU module, may not be fully captured or explicitly present in the output triples from the LU module. Therefore, our approach of double fine-tuning aims to leverage the model's internalized understanding of context and rules. Further, to check our hypothesis, experiments are conducted utilizing LU triples and compared with our proposed \textsc{MedLogic-AQA} model. The results of these experiments provide valuable insights into the effectiveness of the model when trained with externally obtained triples versus internally generated ones and are shown in Tables \ref{tab:bioasq_exp} and \ref{tab:mashqa_exp}.

\begin{table*}[hbt!]\footnotesize
\centering
\begin{tabular}{l|ccccccc}
\hline
\textbf{Models} &  \textbf{Medical Entity F1\% } &  \textbf{BLEU} &  \textbf{ROUGE-L } & \textbf{METEOR} & \textbf{Embedding Average} & \textbf{A-LEN} \\
\hline
\text{LLama2+Rule +Triples}  & 27.12 & 0.2511 & 0.2678 & 0.4301 & 0.827 & 65.51   \\
\hline
\textsc{\textbf{MedLogic-AQA}} & \textbf{38.47} & \textbf{0.2729} &  \textbf{0.2768} & \textbf{0.4383} & \textbf{0.838} & \textbf{53.71}   \\
\hline
\end{tabular}
\caption{\label{tab:bioasq_exp}{Experimental results comparing the performance on the BioASQ dataset.}}
\end{table*}

\begin{table*}[hbt!]\footnotesize
\centering
\begin{tabular}{l|ccccccc}
\hline
\textbf{Models} &  \textbf{Medical Entity F1\% } &  \textbf{BLEU} &  \textbf{ROUGE-L } & \textbf{METEOR} & \textbf{Embedding Average} & \textbf{A-LEN} \\
\hline
\text{LLama2+Rule +Triples}  & 28.48 & 0.2115 & 0.2211 & 0.1971 & 0.761 & 91.47   \\
\hline
\textsc{\textbf{MedLogic-AQA}} & \textbf{31.87} & \textbf{0.2284} &  \textbf{0.2513} & \textbf{0.1969} & \textbf{0.788} & \textbf{56.47}   \\
\hline
\end{tabular}
\caption{\label{tab:mashqa_exp}{Experimental results comparing the performance on the MASHQA dataset.}}
\end{table*}

\newpage
\section*{Frequently Asked Questions (FAQ)}

\subsection*{* How does MedLogic-AQA handle the inherent complexities and nuances of medical terminology and contexts, particularly in generating logically coherent answers?}
\textrightarrow MedLogic-AQA addresses the complexities of medical terminology and contexts through the integration of first-order logic-based rules extracted from medical data sources like UMLS. These rules help the system discern complex logical structures and relationships within medical contexts, enabling it to generate answers that are logically coherent and contextually relevant.

\subsection*{* What distinguishes the Logic Understanding (LU) module from traditional natural language understanding models in the context of MedLogic-AQA?}
\textrightarrow The LU module in MedLogic-AQA differs from traditional natural language understanding models by its focus on extracting logical relationships and rules from medical contexts and questions. While traditional models may prioritize semantic understanding, the LU model emphasizes the identification of first-order logic-based rules and associations, enabling more nuanced reasoning and inference in medical question answering.

\subsection*{* What are the limitations and potential areas for improvement in the LU model, and how might future research address these challenges?}
\textrightarrow Some limitations of the LU model include its reliance on pre-existing knowledge graphs and datasets, which may limit coverage and relevance, and its susceptibility to noise and inaccuracies in entity recognition and relation extraction. Future research could focus on enhancing the robustness and adaptability of the model through improved data preprocessing techniques, more sophisticated logic rule extraction algorithms, and integration with external knowledge sources to enrich the representation of medical concepts and relationships.

\subsection*{* How are logical rules derived and selected for integration into MedLogic-AQA, and what criteria are used to determine their relevance and effectiveness?}
\textrightarrow Logical rules in MedLogic-AQA are derived from comprehensive analysis of medical literature, ontologies, and domain-specific knowledge bases. The selection process involves identifying rules that capture common patterns and relationships within medical data while minimizing redundancy and ambiguity. Criteria for selecting logical rules include their applicability across diverse medical domains, interpretability, and ability to capture nuanced logical dependencies relevant to question answering tasks.

\subsection*{* What are the potential applications of MedLogic-AQA beyond medical question-answering, and how might it contribute to advancements in healthcare technology and research?}
\textrightarrow Beyond medical question-answering, MedLogic-AQA holds potential applications in clinical decision support systems, medical education, and biomedical research. By providing accurate and contextually relevant answers to complex medical queries, the system can assist healthcare professionals in making informed decisions, facilitate medical education and training, and contribute to the discovery of new insights and knowledge in healthcare and biomedicine.


\begin{thebibliography}{}

\bibitem[{Amador-Dom{\'\i}nguez et~al.(2023)Amador-Dom{\'\i}nguez, Serrano, and Manrique}]{amador2023geni}
Elvira Amador-Dom{\'\i}nguez, Emilio Serrano, and Daniel Manrique. 2023.
\newblock Geni: A framework for the generation of explanations and insights of knowledge graph embedding predictions.
\newblock \emph{Neurocomputing}, 521:199--212.

\bibitem[{Asai and Hajishirzi(2020)}]{asai2020logic}
Akari Asai and Hannaneh Hajishirzi. 2020.
\newblock Logic-guided data augmentation and regularization for consistent question answering.
\newblock \emph{arXiv preprint arXiv:2004.10157}.

\bibitem[{Banerjee and Lavie(2005)}]{banerjee2005meteor}
Satanjeev Banerjee and Alon Lavie. 2005.
\newblock Meteor: An automatic metric for mt evaluation with improved correlation with human judgments.
\newblock In \emph{Proceedings of the acl workshop on intrinsic and extrinsic evaluation measures for machine translation and/or summarization}, pages 65--72.

\bibitem[{Berant and Liang(2014)}]{berant2014semantic}
Jonathan Berant and Percy Liang. 2014.
\newblock Semantic parsing via paraphrasing.
\newblock In \emph{Proceedings of the 52nd Annual Meeting of the Association for Computational Linguistics (Volume 1: Long Papers)}, pages 1415--1425.

\bibitem[{Bodenreider(2004)}]{bodenreider2004unified}
Olivier Bodenreider. 2004.
\newblock The unified medical language system (umls): integrating biomedical terminology.
\newblock \emph{Nucleic acids research}, 32(suppl\_1):D267--D270.

\bibitem[{Cappanera(2023)}]{cappanera2023logic}
Paolo Cappanera. 2023.
\newblock Logic in computer science.
\newblock \emph{arXiv preprint arXiv:2301.02454}.

\bibitem[{Choi et~al.(2017)Choi, Schuetz, Stewart, and Sun}]{choi2017doctor}
Edward Choi, Andy Schuetz, Walter~F Stewart, and Jimeng Sun. 2017.
\newblock Doctor ai: Predicting clinical events via recurrent neural networks.
\newblock \emph{arXiv preprint arXiv:1511.05942}.

\bibitem[{Cohen(1960)}]{cohen1960coefficient}
Jacob Cohen. 1960.
\newblock A coefficient of agreement for nominal scales.
\newblock \emph{Educational and psychological measurement}, 20(1):37--46.

\bibitem[{Fan et~al.(2019)Fan, Jernite, Perez, Grangier, Weston, and Auli}]{fan2019eli5}
Angela Fan, Yacine Jernite, Ethan Perez, David Grangier, Jason Weston, and Michael Auli. 2019.
\newblock Eli5: Long form question answering.
\newblock In \emph{Proceedings of the 57th Annual Meeting of the Association for Computational Linguistics}, pages 3558--3567.

\bibitem[{Fouladvand et~al.(2023)Fouladvand, Gomez, Nilforoshan, Schwede, Noshad, Jee, You, Leskovec, Chen et~al.}]{fouladvand2023graph}
Sajjad Fouladvand, Federico~Reyes Gomez, Hamed Nilforoshan, Matthew Schwede, Morteza Noshad, Olivia Jee, Jiaxuan You, Jure Leskovec, Jonathan Chen, et~al. 2023.
\newblock Graph-based clinical recommender: Predicting specialists procedure orders using graph representation learning.
\newblock \emph{Journal of Biomedical Informatics}, page 104407.

\bibitem[{Huai et~al.(2023)Huai, Yang, Tao et~al.}]{huai2023spatial}
Zepeng Huai, Guohua Yang, Jianhua Tao, et~al. 2023.
\newblock Spatial-temporal knowledge graph network for event prediction.
\newblock \emph{Neurocomputing}, page 126557.

\bibitem[{Huth and Ryan(2004)}]{huth2004logic}
Michael Huth and Mark Ryan. 2004.
\newblock \emph{Logic in Computer Science: Modelling and reasoning about systems}.
\newblock Cambridge university press.

\bibitem[{Krishna et~al.(2021)Krishna, Roy, and Iyyer}]{krishna2021hurdles}
Kalpesh Krishna, Aurko Roy, and Mohit Iyyer. 2021.
\newblock Hurdles to progress in long-form question answering.
\newblock In \emph{Proceedings of the 2021 Conference of the North American Chapter of the Association for Computational Linguistics: Human Language Technologies}, pages 4940--4957.

\bibitem[{Labrak et~al.(2024)Labrak, Bazoge, Morin, Gourraud, Rouvier, and Dufour}]{labrak2024biomistral}
Yanis Labrak, Adrien Bazoge, Emmanuel Morin, Pierre-Antoine Gourraud, Mickael Rouvier, and Richard Dufour. 2024.
\newblock Biomistral: A collection of open-source pretrained large language models for medical domains.
\newblock \emph{arXiv preprint arXiv:2402.10373}.

\bibitem[{Leaman et~al.(2015)Leaman, Khare, and Lu}]{leaman2015challenges}
Robert Leaman, Ritu Khare, and Zhiyong Lu. 2015.
\newblock Challenges in clinical natural language processing for automated disorder normalization.
\newblock \emph{Journal of biomedical informatics}, 57:28--37.

\bibitem[{Lewis et~al.(2020)Lewis, Liu, Goyal, Ghazvininejad, Mohamed, Levy, Stoyanov, and Zettlemoyer}]{lewis2020bart}
Mike Lewis, Yinhan Liu, Naman Goyal, Marjan Ghazvininejad, Abdelrahman Mohamed, Omer Levy, Veselin Stoyanov, and Luke Zettlemoyer. 2020.
\newblock Bart: Denoising sequence-to-sequence pre-training for natural language generation, translation, and comprehension.
\newblock In \emph{Proceedings of the 58th Annual Meeting of the Association for Computational Linguistics}, pages 7871--7880.

\bibitem[{Li and Srikumar(2019)}]{li2019augmenting}
Tao Li and Vivek Srikumar. 2019.
\newblock Augmenting neural networks with first-order logic.
\newblock In \emph{Proceedings of the 57th Annual Meeting of the Association for Computational Linguistics}, pages 292--302.

\bibitem[{Li et~al.(2023)Li, Bubeck, Eldan, Del~Giorno, Gunasekar, and Lee}]{li2023textbooks}
Yuanzhi Li, S{\'e}bastien Bubeck, Ronen Eldan, Allie Del~Giorno, Suriya Gunasekar, and Yin~Tat Lee. 2023.
\newblock Textbooks are all you need ii: phi-1.5 technical report.
\newblock \emph{arXiv preprint arXiv:2309.05463}.

\bibitem[{Lin(2004)}]{lin-2004-rouge}
Chin-Yew Lin. 2004.
\newblock \href {https://www.aclweb.org/anthology/W04-1013} {{ROUGE}: A package for automatic evaluation of summaries}.
\newblock In \emph{Text Summarization Branches Out}, pages 74--81, Barcelona, Spain. Association for Computational Linguistics.

\bibitem[{Lin et~al.(2022)Lin, Quan, Wang, Guo, Zeng, and Philip}]{lin2022effectively}
Xuan Lin, Zhe Quan, Zhi-Jie Wang, Yan Guo, Xiangxiang Zeng, and S~Yu Philip. 2022.
\newblock Effectively identifying compound-protein interaction using graph neural representation.
\newblock \emph{IEEE/ACM Transactions on Computational Biology and Bioinformatics}.

\bibitem[{Liu et~al.(2016)Liu, Lowe, Serban, Noseworthy, Charlin, and Pineau}]{liu2016not}
Chia-Wei Liu, Ryan Lowe, Iulian Serban, Mike Noseworthy, Laurent Charlin, and Joelle Pineau. 2016.
\newblock \href {https://doi.org/10.18653/v1/D16-1230} {How {NOT} to evaluate your dialogue system: An empirical study of unsupervised evaluation metrics for dialogue response generation}.
\newblock In \emph{Proceedings of the 2016 Conference on Empirical Methods in Natural Language Processing}, pages 2122--2132, Austin, Texas. Association for Computational Linguistics.

\bibitem[{Loshchilov and Hutter(2018)}]{loshchilov2018decoupled}
Ilya Loshchilov and Frank Hutter. 2018.
\newblock Decoupled weight decay regularization.
\newblock In \emph{International Conference on Learning Representations}.

\bibitem[{Luo et~al.(2022)Luo, Sun, Xia, Qin, Zhang, Poon, and Liu}]{luo2022biogpt}
Renqian Luo, Liai Sun, Yingce Xia, Tao Qin, Sheng Zhang, Hoifung Poon, and Tie-Yan Liu. 2022.
\newblock Biogpt: generative pre-trained transformer for biomedical text generation and mining.
\newblock \emph{Briefings in bioinformatics}, 23(6):bbac409.

\bibitem[{Luo et~al.(2023)Luo, Zhang, Fan, Yang, Wu, Qiao, and Nie}]{luo2023biomedgpt}
Yizhen Luo, Jiahuan Zhang, Siqi Fan, Kai Yang, Yushuai Wu, Mu~Qiao, and Zaiqing Nie. 2023.
\newblock Biomedgpt: Open multimodal generative pre-trained transformer for biomedicine.
\newblock \emph{arXiv preprint arXiv:2308.09442}.

\bibitem[{Minsky(1975)}]{minsky1975framework}
Marvin Minsky. 1975.
\newblock A framework for representing knowledge.
\newblock \emph{MIT-AI Laboratory Memo}.

\bibitem[{Mishra et~al.(2022)Mishra, Firdaus, and Ekbal}]{mishra2022please}
Kshitij Mishra, Mauajama Firdaus, and Asif Ekbal. 2022.
\newblock Please be polite: Towards building a politeness adaptive dialogue system for goal-oriented conversations.
\newblock \emph{Neurocomputing}, 494:242--254.

\bibitem[{Mishra et~al.(2023{\natexlab{a}})Mishra, Priya, Burja, and Ekbal}]{mishra2023therapist}
Kshitij Mishra, Priyanshu Priya, Manisha Burja, and Asif Ekbal. 2023{\natexlab{a}}.
\newblock e-therapist: I suggest you to cultivate a mindset of positivity and nurture uplifting thoughts.
\newblock In \emph{Proceedings of the 2023 Conference on Empirical Methods in Natural Language Processing}, pages 13952--13967.

\bibitem[{Mishra et~al.(2023{\natexlab{b}})Mishra, Priya, and Ekbal}]{mishra2023help}
Kshitij Mishra, Priyanshu Priya, and Asif Ekbal. 2023{\natexlab{b}}.
\newblock Help me heal: A reinforced polite and empathetic mental health and legal counseling dialogue system for crime victims.
\newblock In \emph{Proceedings of the AAAI Conference on Artificial Intelligence}, volume~37, pages 14408--14416.

\bibitem[{Mishra et~al.(2023{\natexlab{c}})Mishra, Priya, and Ekbal}]{mishra2023pal}
Kshitij Mishra, Priyanshu Priya, and Asif Ekbal. 2023{\natexlab{c}}.
\newblock Pal to lend a helping hand: Towards building an emotion adaptive polite and empathetic counseling conversational agent.
\newblock In \emph{Proceedings of the 61st Annual Meeting of the Association for Computational Linguistics (Volume 1: Long Papers)}, pages 12254--12271.

\bibitem[{Moldovan et~al.(2003)Moldovan, Clark, Harabagiu, and Maiorano}]{moldovan2003cogex}
Dan Moldovan, Chris Clark, Sanda Harabagiu, and Steven~J Maiorano. 2003.
\newblock Cogex: A logic prover for question answering.
\newblock In \emph{Proceedings of the 2003 Human Language Technology Conference of the North American Chapter of the Association for Computational Linguistics}, pages 166--172.

\bibitem[{Nentidis et~al.(2022)Nentidis, Katsimpras, Vandorou, Krithara, Miranda-Escalada, Gasco, Krallinger, and Paliouras}]{nentidis2022overview}
Anastasios Nentidis, Georgios Katsimpras, Eirini Vandorou, Anastasia Krithara, Antonio Miranda-Escalada, Luis Gasco, Martin Krallinger, and Georgios Paliouras. 2022.
\newblock Overview of bioasq 2022: The tenth bioasq challenge on large-scale biomedical semantic indexing and question answering.
\newblock In \emph{International Conference of the Cross-Language Evaluation Forum for European Languages}, pages 337--361. Springer.

\bibitem[{Pal et~al.(2022)Pal, Kanoulas, and Rijke}]{pal2022parameter}
Vaishali Pal, Evangelos Kanoulas, and Maarten Rijke. 2022.
\newblock Parameter-efficient abstractive question answering over tables or text.
\newblock In \emph{Proceedings of the Second DialDoc Workshop on Document-grounded Dialogue and Conversational Question Answering}, pages 41--53.

\bibitem[{Papineni et~al.(2002)Papineni, Roukos, Ward, and Zhu}]{papineni2002bleu}
Kishore Papineni, Salim Roukos, Todd Ward, and Wei-Jing Zhu. 2002.
\newblock Bleu: a method for automatic evaluation of machine translation.
\newblock In \emph{Proceedings of the 40th annual meeting on association for computational linguistics}, pages 311--318. Association for Computational Linguistics.

\bibitem[{Pons et~al.(2016)Pons, Braun, Hunink, and Kors}]{pons2016natural}
Ewoud Pons, Loes~M Braun, Myriam~G Hunink, and Jan~A Kors. 2016.
\newblock Natural language processing in radiology: A systematic review.
\newblock \emph{Radiology}, 279(2):329--343.

\bibitem[{Priya et~al.(2023)Priya, Mishra, Totala, and Ekbal}]{priya2023partner}
Priyanshu Priya, Kshitij Mishra, Palak Totala, and Asif Ekbal. 2023.
\newblock Partner: A persuasive mental health and legal counselling dialogue system for women and children crime victims.
\newblock In \emph{IJCAI}, pages 6183--6191.

\bibitem[{Radford et~al.(2019)Radford, Wu, Child, Luan, Amodei, Sutskever et~al.}]{radford2019language}
Alec Radford, Jeffrey Wu, Rewon Child, David Luan, Dario Amodei, Ilya Sutskever, et~al. 2019.
\newblock Language models are unsupervised multitask learners.
\newblock \emph{OpenAI blog}, 1(8):9.

\bibitem[{Rajpurkar et~al.(2018)Rajpurkar, Zhang, Lopyrev, and Liang}]{rajpurkar2018squad}
Pranav Rajpurkar, Jian Zhang, Konstantin Lopyrev, and Percy Liang. 2018.
\newblock Squad: 100,000+ questions for machine comprehension of text.
\newblock In \emph{Proceedings of the 2016 Conference on Empirical Methods in Natural Language Processing}.

\bibitem[{Ratner et~al.(2017)Ratner, Bach, Ehrenberg, Fries, Wu, and R{\'e}}]{ratner2017snorkel}
Alexander~J Ratner, Stephen~H Bach, Henry Ehrenberg, Jason Fries, Sen Wu, and Christopher R{\'e}. 2017.
\newblock Snorkel: Rapid training data creation with weak supervision.
\newblock \emph{Proceedings of the VLDB Endowment}, 11(3):269--282.

\bibitem[{Samad et~al.(2022)Samad, Mishra, Firdaus, and Ekbal}]{samad2022empathetic}
Azlaan~Mustafa Samad, Kshitij Mishra, Mauajama Firdaus, and Asif Ekbal. 2022.
\newblock Empathetic persuasion: reinforcing empathy and persuasiveness in dialogue systems.
\newblock In \emph{Findings of the Association for Computational Linguistics: NAACL 2022}, pages 844--856.

\bibitem[{Shickel et~al.(2018)Shickel, Tighe, Bihorac, and Rashidi}]{shickel2018deep}
Benjamin Shickel, Patrick~J Tighe, Azra Bihorac, and Parisa Rashidi. 2018.
\newblock Deep ehr: A survey of recent advances in deep learning techniques for electronic health record (ehr) analysis.
\newblock \emph{IEEE Journal of Biomedical and Health Informatics}, 22(5):1589--1604.

\bibitem[{Soldaini and Goharian(2016)}]{soldaini2016quickumls}
Luca Soldaini and Nazli Goharian. 2016.
\newblock Quickumls: a fast, unsupervised approach for medical concept extraction.
\newblock In \emph{MedIR workshop, sigir}, pages 1--4.

\bibitem[{Sutskever et~al.(2014)Sutskever, Vinyals, and Le}]{sutskever2014sequence}
Ilya Sutskever, Oriol Vinyals, and Quoc~V Le. 2014.
\newblock Sequence to sequence learning with neural networks.
\newblock \emph{Advances in neural information processing systems}, 27.

\bibitem[{Touvron et~al.(2023)Touvron, Martin, Stone, Albert, Almahairi, Babaei, Bashlykov, Batra, Bhargava, Bhosale et~al.}]{touvron2023llama}
Hugo Touvron, Louis Martin, Kevin Stone, Peter Albert, Amjad Almahairi, Yasmine Babaei, Nikolay Bashlykov, Soumya Batra, Prajjwal Bhargava, Shruti Bhosale, et~al. 2023.
\newblock Llama 2: Open foundation and fine-tuned chat models.
\newblock \emph{arXiv preprint arXiv:2307.09288}.

\bibitem[{Varshney et~al.(2022)Varshney, Zafar, Behera, and Ekbal}]{varshney2022cdialog}
Deeksha Varshney, Aizan Zafar, Niranshu Behera, and Asif Ekbal. 2022.
\newblock Cdialog: A multi-turn covid-19 conversation dataset for entity-aware dialog generation.
\newblock In \emph{Proceedings of the 2022 Conference on Empirical Methods in Natural Language Processing}, pages 11373--11385.

\bibitem[{Varshney et~al.(2023)Varshney, Zafar, Behera, and Ekbal}]{varshney2023knowledge}
Deeksha Varshney, Aizan Zafar, Niranshu~Kumar Behera, and Asif Ekbal. 2023.
\newblock Knowledge graph assisted end-to-end medical dialog generation.
\newblock \emph{Artificial Intelligence in Medicine}, 139:102535.

\bibitem[{Vaswani et~al.(2017)Vaswani, Shazeer, Parmar, Uszkoreit, Jones, Gomez, Kaiser, and Polosukhin}]{vaswani2017attention}
Ashish Vaswani, Noam Shazeer, Niki Parmar, Jakob Uszkoreit, Llion Jones, Aidan~N Gomez, Łukasz Kaiser, and Illia Polosukhin. 2017.
\newblock Attention is all you need.
\newblock In \emph{Advances in neural information processing systems}, pages 5998--6008.

\bibitem[{Wang et~al.(2021)Wang, Sun, Zou, Wu, and Han}]{wang2021logic}
Xinyu Wang, Tao Sun, Deqing Zou, Wei Wu, and Jiawei Han. 2021.
\newblock Logic-guided data augmentation and regularization for consistency learning.
\newblock \emph{arXiv preprint arXiv:2104.04379}.

\bibitem[{Zafar et~al.(2023)Zafar, Sahoo, Bhardawaj, Das, and Ekbal}]{zafar2023ki}
Aizan Zafar, Sovan~Kumar Sahoo, Harsh Bhardawaj, Amitava Das, and Asif Ekbal. 2023.
\newblock Ki-mag: A knowledge-infused abstractive question answering system in medical domain.
\newblock \emph{Neurocomputing}, page 127141.

\bibitem[{Zafar et~al.(2024{\natexlab{a}})Zafar, Sahoo, Varshney, Das, and Ekbal}]{zafar2024kimedqa}
Aizan Zafar, Sovan~Kumar Sahoo, Deeksha Varshney, Amitava Das, and Asif Ekbal. 2024{\natexlab{a}}.
\newblock Kimedqa: towards building knowledge-enhanced medical qa models.
\newblock \emph{Journal of Intelligent Information Systems}, pages 1--26.

\bibitem[{Zafar et~al.(2024{\natexlab{b}})Zafar, Varshney, Kumar~Sahoo, Das, and Ekbal}]{zafar2024my}
Aizan Zafar, Deeksha Varshney, Sovan Kumar~Sahoo, Amitava Das, and Asif Ekbal. 2024{\natexlab{b}}.
\newblock Are my answers medically accurate? exploiting medical knowledge graphs for medical question answering.
\newblock \emph{Applied Intelligence}, 54(2):2172--2187.

\bibitem[{Zhang et~al.(2020)Zhang, Zhang, Xia, and Sun}]{zhang2020graph}
Jiawei Zhang, Haopeng Zhang, Congying Xia, and Li~Sun. 2020.
\newblock Graph-bert: Only attention is needed for learning graph representations.
\newblock \emph{arXiv preprint arXiv:2001.05140}.

\bibitem[{Zhu et~al.(2020{\natexlab{a}})Zhu, Xiong, and Socher}]{zhu2020knowledge}
Jie Zhu, Chenyan Xiong, and Richard Socher. 2020{\natexlab{a}}.
\newblock Knowledge-driven semantic role labeling: A new perspective for interpreting human activity.
\newblock \emph{arXiv preprint arXiv:2011.06745}.

\bibitem[{Zhu et~al.(2020{\natexlab{b}})Zhu, Ahuja, Juan, Wei, and Reddy}]{zhu2020question}
Ming Zhu, Aman Ahuja, Da-Cheng Juan, Wei Wei, and Chandan~K Reddy. 2020{\natexlab{b}}.
\newblock Question answering with long multiple-span answers.
\newblock In \emph{Findings of the Association for Computational Linguistics: EMNLP 2020}, pages 3840--3849.

\end{thebibliography}
\end{document}